%% file: acl_latex.tex
\documentclass[11pt]{article}

\usepackage[final]{acl}
\usepackage{enumitem}

\usepackage{times}
\usepackage{latexsym}
\usepackage{amsmath}   
\usepackage{multirow}
\usepackage{booktabs}
\usepackage{algorithm}
\usepackage{algpseudocode}
\usepackage{amssymb}
\usepackage[T1]{fontenc}

\usepackage[utf8]{inputenc}

\usepackage{microtype}

\usepackage{cuted}
\usepackage{amsthm}
\usepackage{inconsolata}
\usepackage[most]{tcolorbox}
\usepackage{lipsum}
\usepackage{graphicx}
\usepackage{xcolor}
\usepackage[normalem]{ulem}
\useunder{\uline}{\ul}{}

\title{Adaptive Milestone Reward for GUI Agents}

\author{
 \textbf{Congmin Zheng\textsuperscript{1}\thanks{These authors contributed equally.}}\thanks{This work was done during Congmin Zheng’s internship at
OPPO Research Institute.},
 \textbf{Xiaoyun Mo\textsuperscript{2}$^{\ast}$},
 \textbf{Xinbei Ma\textsuperscript{1}},
 \textbf{Qiqiang Lin\textsuperscript{2}},
 \textbf{Yin Zhao\textsuperscript{2}},
 \\
 \textbf{Jiachen Zhu\textsuperscript{1}},
 \textbf{Xingyu Lou\textsuperscript{2}}$^\ddagger$,
 \textbf{Jun Wang \textsuperscript{2}}$^\ddagger$,
 \textbf{Zhaoxiang Wang\textsuperscript{2}},
 \\
 \textbf{Weiwen Liu\textsuperscript{1}}$^\ddagger$,
 \textbf{Zhuosheng Zhang\textsuperscript{1}},
 \textbf{Yong Yu\textsuperscript{1}},
 \textbf{Weinan Zhang\textsuperscript{1}}\thanks{Corresponding author},
\\
 \textsuperscript{1}Shanghai Jiao Tong University,
 \textsuperscript{2}OPPO Research Institute,
\\
\{desp.zcm, wwliu, wnzhang\}@sjtu.edu.cn, junwang.lu@gmail.com\\
\{moxiaoyun, louxingyu\}@oppo.com
}

\begin{document}
\maketitle
\begin{abstract}
Reinforcement Learning (RL) has emerged as a mainstream paradigm for training Mobile GUI Agents, yet it struggles with the temporal credit assignment problem inherent in long-horizon tasks. A primary challenge lies in the trade-off between reward fidelity and density: outcome reward offers high fidelity but suffers from signal sparsity, while process reward provides dense supervision but remains prone to bias and reward hacking. To resolve this conflict, we propose the \textbf{Ad}aptive \textbf{Mi}lestone \textbf{Re}ward (ADMIRE) mechanism. ADMIRE constructs a verifiable, adaptive reward system by anchoring trajectory to milestones, which are dynamically distilled from successful explorations. Crucially, ADMIRE integrates an asymmetric credit assignment strategy that denoises successful trajectories and scaffolds failed trajectories. Extensive experiments demonstrate that ADMIRE consistently yields over 10\% absolute improvement in success rate across different base models on AndroidWorld. Moreover, the method exhibits robust generalizability, achieving strong performance across diverse RL algorithms and heterogeneous environments such as web navigation and embodied tasks.
\end{abstract}

\section{Introduction}

Mobile GUI Agents~\citep{yan2025step,nguyen2025gui,wu2025gem,wu2025quick,wang2025ui,tang2025magicgui,nong2025craft}, driven by the rapid advances in multimodal large language models (MLLMs)~\citep{bai2025qwen2,wang2025internvl3_5,Qwen3-VL,o3Operator}, are emerging as a powerful paradigm for automating tasks on mobile devices. Currently, the evolution of Mobile GUI Agents stands at a critical juncture, transitioning from executing simple directives to navigating complex, long-horizon tasks~\citep{guo2025atomic,song2025colorbench,kong2025mobileworld}. In this domain, Reinforcement Learning (RL) has become the mainstream engine for training, enabling agents to discover optimal operational paths within intricate interfaces through iterative trial-and-error and continuous interaction~\cite{shi2025mobilegui,luo2025gui,zhang2025agentcpm,lu2025ui}. However, the nature of these interactions inevitably gives rise to the temporal credit assignment problem~\citep{sutton1984temporal,pignatelli2023survey}: the challenge of accurately attributing a final outcome to specific actions within long sequences where feedback is sparse and delayed. 

\begin{figure}[t]
  \centering
  \vspace{-15pt}
  \includegraphics[width=1\linewidth]{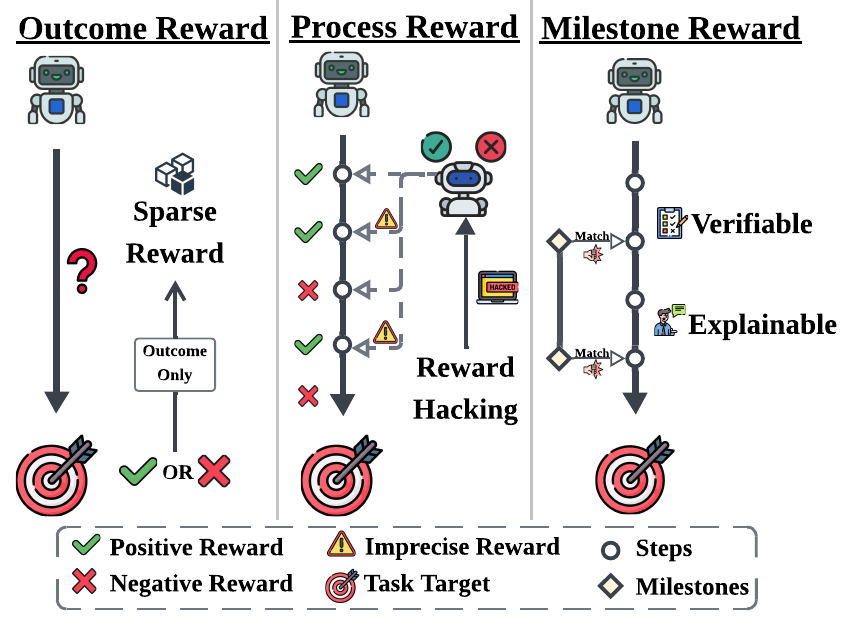}
  \vspace{-20pt}
  \caption{Comparison of different reward mechanisms. Milestones are identified as key state transitions to enable verifiable and interpretable reward triggering.}
  \label{fig: reward comparison}
  \vspace{-15pt}
\end{figure}

However, existing reward mechanisms struggle to adequately address this challenge. As illustrated in Figure~\ref{fig: reward comparison}, outcome reward results in sparse signals, while process reward is vulnerable to reward hacking. Specifically, (i) Outcome reward~\citep{lu2025arpo,xi2025agentgym,xi2025agentprm} evaluates trajectories solely based on the final execution result, typically based on strict rules or system states. While ensuring high fidelity, outcome reward suffers from signal sparsity in long trajectories by reducing complex paths to binary feedback. This simplification prevents the model from recognizing ``near-success'' explorations, thereby hindering efficient exploration. (ii) Process reward~\citep{sun2025seagent,zhang2025progrm,lu2025orcust,dai2025prore} provides dense, step-wise supervision, usually derived from subjective black-box scoring by models. However, this reliance introduces risks of systemic bias and reward hacking~\citep{song2025causal,zheng2025survey}. Critically, process reward fails to differentiate correctness from effectiveness, and rewarding executable but futile actions can trap the agent in suboptimal policies.

This collectively leads to our central research question: \textit{\textbf{How can we achieve dense, step-wise guidance while strictly maintaining the verifiability and fidelity of the reward signal?}} Specifically, we propose the \textbf{Ad}aptive \textbf{Mi}lestone \textbf{Re}ward (ADMIRE) Mechanism. We identify key state transitions within successful explorations to define milestones, which serve as the verifiable basis for triggering rewards via a rule-based matching protocol. Crucially, these milestones are adaptive, dynamically updating to mirror superior behaviors discovered during exploration. This co-evolution aligns the milestone reward with the agent's evolving strategies, ensuring an explainable credit assignment that faithfully reflects genuine task progress. This alignment enables a dense yet principled reward mechanism that continuously adapts to the agent’s growing capabilities.

To maximize signal utility, we incorporate asymmetric credit assignment to effectively leverage both positive and negative trajectories. For successful trajectories, we restrict positive incentives to milestone-triggering steps, compelling the model to distill essential decision points from redundant steps and effectively filtering out process noise. Conversely, for failed trajectories, we apply a dense reward strategy that assigns partial credit via intermediate milestones, constructing a guidance scaffold that breaks the ``all-or-nothing'' paradigm and significantly lowers the barrier to exploration.
%

We conducted extensive experiments on AndroidWorld, which demonstrate that ADMIRE robustly yields over 10\% improvement in success rate across different base models. Furthermore, ADMIRE exhibits strong generalizability, delivering significant performance gains across diverse reinforcement learning algorithms (e.g., GRPO~\citep{shao2024deepseekmath}, RLOO~\citep{ahmadian2024back} and DAPO~\citep{yu2025dapo}) and cross-domain environments, including ALFWorld~\citep{shridhar2020alfworld} and WebShop~\citep{yao2022webshop}.


In summary, our contributions are as follows:

\begin{itemize}[leftmargin=10pt, itemsep=0pt, parsep=0pt, topsep=0pt]
    \vspace{-3pt}
    \item We introduce the Adaptive Milestone Reward mechanism, a novel paradigm that integrates  adaptive, verifiable milestones into online reinforcement learning to provide dense, high-confidence feedback.
    \item We introduce an asymmetric strategy that maximizes the utility of diverse trajectories, enabling robust credit assignment and efficient learning in long-horizon scenarios.
    \item Extensive experiments on AndroidWorld, MobileMiniWob++, and cross-domain tasks demonstrate the effectiveness and strong generalization of our approach.
\end{itemize}

\section{Preliminary}
\label{sec:preliminary}

We formulate the mobile GUI task as a Partially Observable Markov Decision Process (POMDP) defined by the tuple: 
\begin{equation}
\mathcal{M} = \langle \mathcal{S}, \mathcal{A}, \Omega, \mathcal{P}, \mathcal{Z}, \mathcal{R} \rangle.
\end{equation}

The state space $\mathcal{S}$ represents the underlying system status (e.g., app internal states), which is not fully accessible to the agent. Instead, the agent perceives the environment through the observation space $\Omega$. At each time step $t$, the agent receives an observation $o_t$, defined as $o_t = \{ G, I_t \}$, where $G$ is the high-level instruction and $I_t$ is the current interface screenshot. To mitigate partial observability, the agent also references the interaction history $h_t$ to make decisions:
\begin{equation}
h_t = \big( (o_0, a_0), (o_1, a_1), \dots, (o_{t-1}, a_{t-1}), o_t \big).
\end{equation}

The action space $\mathcal{A}$ consists of interface operations available to the agent, such as clicking a button, swiping up. The agent policy $\pi_{\text{agent}}$ maps the interaction history to an action distribution:
\begin{equation}
a_t \sim \pi_{\text{agent}}(a \mid h_t).
\end{equation}

After executing $a_t$, the environment transitions from the latent state $s_t$ to $s_{t+1}$ according to the transition function $\mathcal{P}(s_{t+1} \mid s_t, a_t)$, and subsequently emits a new observation $o_{t+1}$ based on the observation function:
\begin{equation}
o_{t+1} \sim \mathcal{Z}(o \mid s_{t+1}, a_t).
\end{equation}
The agent iteratively repeats this process until the task is completed or a pre-defined maximum step limit is reached. Denoting this terminal step as $T$, the entire interaction process forms a complete trajectory $\tau$, defined as the accumulated history at the end of the episode:
\begin{equation}
\tau = h_T = \big( (o_0, a_0), \dots, (o_{T-1}, a_{T-1}), o_T \big).
\end{equation}

$\mathcal{R}$ denotes the reward function evaluating the agent's performance. Each trajectory is associated with a binary outcome score $\mathcal{O}(\tau) \in \{0, 1\}$, where $1$ signifies success and $0$ indicates failure.

\section{Methodology}
\label{sec:method}

\begin{figure*}
    \centering
    \vspace{-20pt}
    \includegraphics[width=\textwidth]{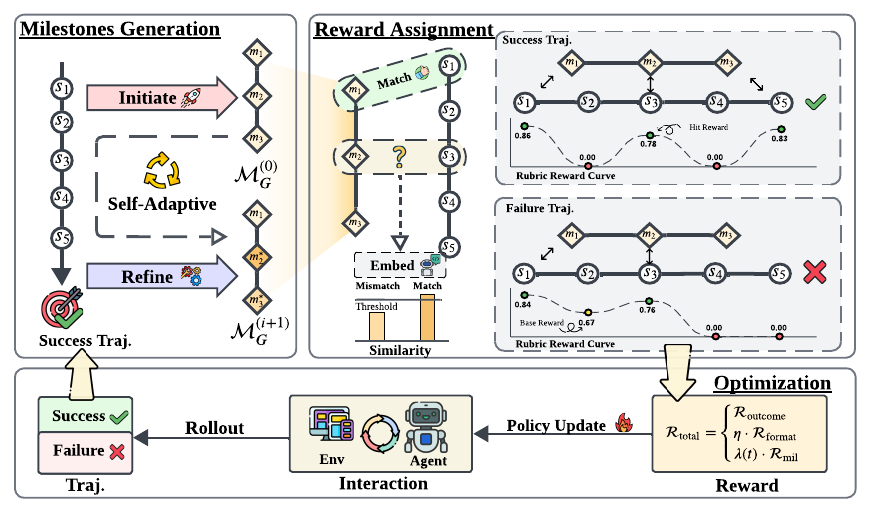}
    \vspace{-15pt}
    \caption{The overall framework of ADMIRE. Circles $s_i$ represent trajectory steps, while diamond markers $m_i$ denote task milestones.}
    \vspace{-20pt}
    \label{fig:rubric_framework}
\end{figure*}
In this section, we present the ADMIRE mechanism. This approach addresses the credit assignment challenges in long-horizon mobile GUI tasks by constructing an objective, adaptive reward system. As illustrated in Figure~\ref{fig:rubric_framework}, the framework consists of three core components: (i) Adaptive Milestone Generation; (ii) Reward Assignment; and (iii) Policy Optimization.

\subsection{Adaptive Milestone Generation}
\label{subsec:milestone_gen}

To transform a specific high-level instruction $G$ into verifiable intermediate signals, we introduce milestones $\mathcal{M}_G$. Unlike static sub-goals, our milestones are dynamically distilled from successful environmental interactions observed during the online training process.

In each training iteration, let $\mathcal{B}$ denote the batch of trajectories collected across multiple tasks under the current policy. For a specific instruction $G$, we filter the subset of successful trajectories, denoted as $\mathcal{B}_G^+ = \{ \tau \in \mathcal{B} \mid \text{Target}(\tau)=G, \mathcal{O}(\tau) = 1 \}$. To acquire the initial milestones $\mathcal{M}_G^{(0)}$, an exemplar trajectory $\tau^* \in \mathcal{B}_G^+$ is processed via a generative abstraction function $\Phi$ (parameterized by a large language model), conditioned on an initialization prompt $\mathbf{P}_{\text{init}}$:
\begin{equation}
    \mathcal{M}_G^{(0)} = \Phi(\tau^*, G, \mathbf{P}_{\text{init}})= [m_1, \dots, m_K].
\end{equation}
Here, $\mathbf{P}_{\text{init}}$ guides the model to abstract critical checkpoints (e.g., ``\textit{Search button clicked}'') from the trajectory $\tau^*$.

Crucially, these milestones are designed to be adaptive rather than static, co-evolving alongside the agent's policy to accommodate emerging superior strategies. For instance, if an agent learns to replace redundant scrolling with a search bar shortcut, the milestones must dynamically update to target this optimized path, ensuring the reward signal correctly reinforces the efficiency.


Let $\mathcal{M}_G^{(i)}$ denote the milestones at iteration $i$. When a new successful trajectory $\tau_{\text{new}}$ is discovered (e.g., finding a shortcut), we invoke the update process using a specific refinement prompt $\mathbf{P}_{\text{update}}$:
\begin{equation}
    \mathcal{M}_G^{(i+1)} = \Phi(\tau_{\text{new}}, \mathcal{M}_G^{(i)}, G , \mathbf{P}_{\text{update}}).
\end{equation}
In this step, $\mathbf{P}_{\text{update}}$ instructs $\Phi$ to compare the new trajectory $\tau_{\text{new}}$ against the existing milestones. If $\tau_{\text{new}}$ represents a more optimal path, the milestones are refined to align with this superior strategy.

\subsection{Reward Assignment}
\subsubsection{Semantic Matching and Verification}
\label{subsec:matching}

To compute the milestone reward, we employ a semantic matching protocol that maps the agent's action to the expected milestone. We utilize a pre-trained Sentence-BERT~\citep{reimers2019sentence} encoder $\psi(\cdot)$ to project textual descriptions into a shared embedding space. At step $t$, we verify the current action description $a_t$ against a candidate milestone $m_k$ by computing their semantic cosine similarity $s(a_t, m_k)$:
\begin{equation}
    s(a_t, m_k) = \frac{\psi(a_t) \cdot \psi(m_k)}{\| \psi(a_t) \| \| \psi(m_k) \|}.
\end{equation}

To enforce sequential constraints, we maintain a pointer $p_t$ tracking the next uncompleted milestone. Consequently, the milestone reward $r^{\text{mil}}_t$ is calculated strictly against the current target $m_{p_t}$:
\begin{equation}
    r^{\text{mil}}_t = \mathbb{I}\left( s(a_t, m_{p_t}) > \delta \right) \cdot s(a_t, m_{p_t}),
\end{equation}
where $\delta$ is a confidence threshold and $\mathbb{I}(\cdot)$ is the indicator function. The pointer $p_t$ increments to $p_t+1$ only if this match condition is met($s(a_t, m_{p_t}) > \delta$), preventing out-of-order skipping.

\subsubsection{Asymmetric Credit Assignment}
\label{subsec:asymmetric}

The crux of our approach lies in reconciling the conflict between maintaining a high fidelity and providing high-frequency feedback. To address this, we differentiate the calculation of the milestone reward $\mathcal{R}_{\text{mil}}$ based on the binary outcome score of the trajectory $\mathcal{O}(\tau) \in \{0, 1\}$. Note that $r^{\text{mil}}_t$ denotes the raw semantic similarity score at step $t$ (as defined in Sec.~\ref{subsec:matching}).

\textbf{Case 1: Denoising Positive Samples ($\mathcal{O}(\tau)=1$).}
Successful episodes often contain redundant operations and noise. To strictly reinforce critical decisions, we apply a denoising mask to solely activate the reward at steps that hit a milestone:
\begin{equation}
    \mathcal{R}_{\text{mil}}(t) = 
    \begin{cases} 
        r^{\text{mil}}_t & \text{if } t \in \mathcal{T}_{\text{mil}}, \\
        0 & \text{otherwise},
    \end{cases}
\end{equation}
where $\mathcal{T}_{\text{mil}}$ denotes the steps that hit a milestone. This filters out non-essential actions, ensuring the policy converges towards the most efficient path.

\textbf{Case 2: Scaffolding Negative Samples ($\mathcal{O}(\tau)=0$).}
Failed trajectories are often underutilized, resulting in wasted exploration. To salvage this signal, we adopt a dense scaffolding reward scheme composed of a base progress reward and a milestone hit bonus. Let $K$ denote the total number of milestones, $k$ the number of milestones achieved within the trajectory and $\zeta$ a fixed weight. The step-wise reward at time $t$ is defined as:
\begin{equation}
    \mathcal{R}_{\text{mil}}(t) = \frac{k}{K} +
    \begin{cases} 
         \zeta \cdot r^{\text{mil}}_t & \text{if } t \in \mathcal{T}_{\text{mil}}, \\
        0& \text{otherwise}.
    \end{cases}
\end{equation}

This mechanism guarantees that the agent receives a continuous base reward ($\frac{k}{K}$) for maintaining progress, supplemented by a bonus for achieving specific milestones. By validating partial successes, we break the ``all-or-nothing'' paradigm and significantly lower the exploration barrier.

\subsection{Policy Optimization}
\label{subsec:grpo}
We integrate ADMIRE into the Group Relative Policy Optimization (GRPO)~\citep{shao2024deepseekmath} and follow \citet{lai2025computerrl} in extending GRPO to the step level to improve training efficiency.

For each instruction $G$ sampled from the dataset $\mathcal{G}$, we generate a group of $B$ trajectories $\{\tau_i\}_{i=1}^B$ under the current policy $\pi_{\theta_{\text{old}}}$. Let $T_i$ denote the terminal step of trajectory $\tau_i$. The trajectory consists of a sequence of pairs $\{(h^i_t, a^i_t)\}_{t=1}^{T_i}$. The optimization objective is formulated as:

\begin{equation}
\label{eq:step_grpo_loss}
\small
\begin{aligned}
    & \mathcal{J}(\theta) = \mathbb{E}_{G \sim \mathcal{G}, \{\tau_i\} \sim \pi_{\theta_{\text{old}}}} \Bigg[ \frac{1}{\sum_{i=1}^{B} T_i} \sum_{i=1}^{B} \sum_{t=1}^{T_i} \min \\
    &\Bigg( \frac{\pi_\theta(a^i_t \mid h^i_t)}{\pi_{\theta_{\text{old}}}(a^i_t \mid h^i_t)} \hat{A}^i_t,  \text{clip}\left(\frac{\pi_\theta(a^i_t \mid h^i_t)}{\pi_{\theta_{\text{old}}}(a^i_t \mid h^i_t)}, 1-\epsilon, 1+\epsilon\right) \hat{A}^i_t \Bigg) \Bigg].
\end{aligned}
\end{equation}

\begin{equation}
\small
    \hat{A}^i_t = \frac{\mathcal{R}^i_{\text{total}}(t) - \text{mean}(\mathbf{R})}{\text{std}(\mathbf{R})},
\end{equation}
where $\mathbf{R} = \{ \mathcal{R}^u_{\text{total}}(v) \mid 1 \le u \le B, 1 \le v \le T_u \}$. The total reward $\mathcal{R}^i_{\text{total}}(t)$ aggregates the outcome success, format validity, and milestone guidance:

Here, $\mathcal{R}^i_{\text{outcome}}$ is the binary task success indicator determined by $\mathcal{O}(\tau) \in \{0, 1\}$; $\mathcal{R}^i_{\text{format}}$ is set to $-1$ if the action syntax is invalid and $0$ otherwise, and $\mathcal{R}^i_{\text{mil}}$ represents the asymmetric reward  derived in Sec.~\ref{subsec:asymmetric}. To balance exploration and exploitation, the curriculum coefficient $ \lambda_0 \cdot \gamma^{\mathcal{E}}$ (where $\mathcal{E}$ is the training epoch) dynamically decays the dense milestone reward over time, gradually shifting the optimization focus toward the outcome reward.

The complete training procedure of ADMIRE is formally presented in Algorithm~\ref{alg:admire}.

\section{Experiment}

\begin{table*}[t]
\centering\small
\vspace{-15pt}
\setlength{\tabcolsep}{10pt} 
{
\begin{tabular}{cccccccc}
\toprule
                                                              &                                 & \multicolumn{2}{c}{AndroidWorld}      & \multicolumn{2}{c}{MobileMiniWob++}   & \multicolumn{2}{c}{Avg} \\ \cmidrule(r){3-4} \cmidrule(r){5-6} \cmidrule(r){7-8} 
\multirow{-2}{*}{Base Model}                                  & \multirow{-2}{*}{Reward Source} & SR (\%) & $\Delta$                    & SR (\%) & $\Delta$                    & SR (\%)    & $\Delta$   \\ \midrule
\multicolumn{1}{c|}{}                                         & -                               & 18.1    & -                           & 52.1    & -                           & 35.1       & -          \\
\multicolumn{1}{c|}{}                                         & Outcome Reward                             & 26.7    & {\color[HTML]{32CB00} +8.6}  & \underline{55.6}    & {\color[HTML]{32CB00} +3.5}  & \underline{41.2}       & {\color[HTML]{32CB00} +6.1} \\
\multicolumn{1}{c|}{}                                         & Process Reward                             & \underline{27.5}    & {\color[HTML]{32CB00} +9.4}  & 54.3    & {\color[HTML]{32CB00} +2.2}  & 40.9       & {\color[HTML]{32CB00} +5.8} \\
\multicolumn{1}{c|}{\multirow{-4}{*}{Qwen2.5-VL-3B-Instruct}} & ADMIRE(Ours)                            & \textbf{31.0$^*$}    & {\color[HTML]{32CB00} +12.9} & \textbf{57.6$^*$}    & {\color[HTML]{32CB00} +5.5}  & \textbf{44.3}       & {\color[HTML]{32CB00} +9.2} \\ \midrule
\multicolumn{1}{c|}{}                                         & -                               & 32.8    & -                           & 57.6    & -                           & 45.2       & -          \\
\multicolumn{1}{c|}{}                                         & Outcome Reward                             & \underline{39.7}    & {\color[HTML]{32CB00} +6.9}  & 51.1    & {\color[HTML]{FE0000} -6.5} & 45.4       & {\color[HTML]{32CB00} +0.2} \\
\multicolumn{1}{c|}{}                                         & Process Reward                             & 36.2    & {\color[HTML]{32CB00} +3.4}  & \underline{58.8}    & {\color[HTML]{32CB00} +1.2}  & \underline{47.5}       & {\color[HTML]{32CB00} +2.3} \\
\multicolumn{1}{c|}{\multirow{-4}{*}{Qwen2.5-VL-7B-Instruct}} & ADMIRE(Ours)                            & \textbf{44.0$^*$}    & {\color[HTML]{32CB00} +11.2} & \textbf{61.1$^*$}    & {\color[HTML]{32CB00} +3.5}  & \textbf{52.6}       & {\color[HTML]{32CB00} +7.4} \\ \bottomrule
\end{tabular}
}
\vspace{-5pt}
\caption{Comparative performance of models trained with ADMIRE, Outcome Reward, and Process Reward on AndroidWorld and MobileMiniWob++. $\Delta$ denotes the performance gap between the trained model and the corresponding base model. The best result is given in \textbf{bold}, and the second-best value is \underline{underlined}. $^*$ indicates statistical significance ($p < 0.05$).}
\label{tab:main-reward}
\vspace{-10pt}
\end{table*}



\begin{table}[ht]
\centering
\setlength{\tabcolsep}{2pt} 
\resizebox{1.0\linewidth}{!}{
\begin{tabular}{lcc}
\toprule
Model                & \#Params & SR(\%) \\ \midrule
\textit{Proprietary Models  }       &        &                \\
GPT-4o~\citep{achiam2023gpt}       & -      & 34.5              \\
Claude-Sonnet-4~\citep{claude4}          & -       &  41.0            \\ \midrule
\textit{Open Source Models}       &        &                \\
InfiGUIAgent~\citep{liu2025infiguiagent}         & 2B       & 9.0               \\
OS-Genesis~\citep{sun2025genesis}           & 7B       & 17.4            \\
GUI-PRA~\citep{xiong2025gui}              & 7B       & 21.1            \\
Aguvis~\citep{xu2024aguvis}               & 72B      & 26.1            \\
GUI-Critic-R1~\citep{wanyan2025look}        & 7B       & 27.6            \\
MobileGUI-7B~\citep{shi2025mobilegui}         & 7B       & 30.0              \\
UI-TARS-1.5-7B~\citep{qin2025ui}       & 7B       & 32.8            \\
Qwen2.5-VL-72B~\citep{bai2025qwen2}       & 72B      & 35.0              \\
GUI-Shepherd~\citep{chen2025gui}         & 7B       & 40.5            \\
GLM-4.1V-9B-Thinking~\citep{hong2025glm} & 9B       & {\ul41.7 }           \\ \midrule
\textit{Ours}       &        &                \\
ADMIRE(w/Qwen2.5-VL-3B-Instruct)                 & 3B       & 31.0              \\
ADMIRE(w/Qwen2.5-VL-7B-Instruct)                 & 7B       & \textbf{44.0}              \\ \bottomrule
\end{tabular}
}
\vspace{-5pt}
\caption{Performance Comparison with Baselines on the AndroidWorld Benchmark. The best result is given in \textbf{bold}, and the second-best value is \underline{underlined}.}
\label{tab:baseline}
\vspace{-10pt}
\end{table}

\begin{table*}[ht]
\centering\small
\setlength{\tabcolsep}{10pt} 
\vspace{-12pt}
{
\begin{tabular}{cccccccc}
\toprule
                                                              &                                                                             & \multicolumn{2}{c}{AndroidWorld}        & \multicolumn{2}{c}{MobileMiniWob++}     & \multicolumn{2}{c}{Avg} \\ \cmidrule(r){3-4} \cmidrule(r){5-6} \cmidrule(r){7-8}  
\multirow{-2}{*}{Base Model}                                  & \multirow{-2}{*}{\begin{tabular}[c]{@{}c@{}}Milestone\\ State\end{tabular}} & SR (\%)       & $\Delta$                                   & SR (\%)       & $\Delta$                                   & SR (\%)    & $\Delta$   \\ \midrule
\multicolumn{1}{c|}{}                                         & -                                                                           & 18.1          & -                                          & 52.1          & -                                          & 35.1       & -          \\
\multicolumn{1}{c|}{}                                         & Static(Human)                                                               & {\ul 28.4}    & {\color[HTML]{32CB00} +10.3}                & {\ul 55.6}    & {\color[HTML]{32CB00} +3.5}                 & {\ul 42.0}        & {\color[HTML]{32CB00} +6.9} \\
\multicolumn{1}{c|}{}                                         & Static (7B)                                           & 27.6      & {\color[HTML]{32CB00} +9.5}                 & 54.4          & {\color[HTML]{32CB00} +2.3}                 & 41.0       & {\color[HTML]{32CB00} +5.9} \\
\multicolumn{1}{c|}{\multirow{-4}{*}{Qwen2.5-VL-3B-Instruct}} & Adaptive                                                                     & \textbf{31.0} & {\color[HTML]{32CB00} +12.9}                & \textbf{57.6} & {\color[HTML]{32CB00} +5.5}                 & \textbf{44.3}       & {\color[HTML]{32CB00} +9.2} \\ \midrule
\multicolumn{1}{c|}{}                                         & -                                                                           & 32.8          & -                                          & 57.6          & -                                          & 45.2       & -          \\
\multicolumn{1}{c|}{}                                         & Static (Human)                                                                   & { \ul41.4}    & {\color[HTML]{32CB00} +8.6}                 & \textbf{63.0}   & {\color[HTML]{32CB00} +5.4}                 &{\ul 52.2 }       & {\color[HTML]{32CB00} +7.0} \\
\multicolumn{1}{c|}{\multirow{-3}{*}{Qwen2.5-VL-7B-Instruct}} & Adaptive                                                                     & \textbf{44.0} & {\color[HTML]{32CB00} +11.2}                & {\ul 61.1}    & {\color[HTML]{32CB00} +3.5}                 & \textbf{52.6}       & {\color[HTML]{32CB00} +7.4} \\ \bottomrule
\end{tabular}
}
\vspace{-5pt}
\caption{Comparison of static and Adaptive milestones in reward shaping. Static (Human) uses fixed human-annotated milestones, whereas Static (7B) adopts milestones iteratively updated during 7B training and fixed afterward. $\Delta$ indicates the performance gain over the base model. The best result is given in \textbf{bold}, and the second-best value is \underline{underlined}.}
\label{tab:milestone-compare}
\vspace{-5pt}
\end{table*}

In this section, we present the experimental settings
and overall performance.

\subsection{Experiment Settings}
\subsubsection{Implementation Details}

Our GUI agent is built upon the Qwen2.5-VL~\citep{bai2025qwen2} series, specifically utilizing the 3B and 7B instruction-tuned variants. To facilitate scalable, efficient online trajectory collection, we employ a distributed server architecture comprising a pool of environment workers. Each worker is implemented as an Android Virtual Device (AVD) running the AndroidWorld~\citep{rawles2024androidworld} sandbox. Details for training hyperparameters, infrastructure setup, and prompt designs are provided in Appendix~\ref{app:Implementation Details}.

\subsubsection{Benchmarks and Baselines}
\paragraph{Benchmarks:}We evaluate our agents using Success Rate (SR) across mobile-centric benchmarks AndroidWorld~\citep{rawles2024androidworld} and MobileMiniWob++~\citep{liu2018reinforcement,rawles2024androidworld}. 

\paragraph{Baselines:}We evaluate our method against strong baselines trained with outcome and process reward. Comparisons also include leading proprietary models like GPT-4o~\citep{achiam2023gpt} and prominent open-source models like  UI-TARS-1.5-7B~\citep{bai2025qwen2}, and GLM-4V-9B-Thinking~\citep{hong2025glm}.
Further details regarding the benchmarks and baselines are provided in the Appendix~\ref{app:benchmark baseline}.

\subsection{Overall Performance}
\label{sec:overall_performance}

Tables~\ref{tab:main-reward} and~\ref{tab:baseline} summarize ADMIRE’s main results across benchmarks, from which we draw the following observations:

\begin{itemize}[leftmargin=10pt]
    \vspace{-5pt}
    \item As shown in Table~\ref{tab:main-reward}, ADMIRE consistently outperforms the base models and traditional reward mechanisms (outcome and process reward) across all settings. For Qwen2.5-VL-3B, ADMIRE yields a 9.2\% average success rate gain, outperforming outcome (6.1\%) and process (5.8\%) reward. Similarly, ADMIRE (w/Qwen2.5-VL-7B) achieves the highest average success rate of 52.6\%, demonstrating its advantage over other reward mechanisms.
    \vspace{-5pt}
    \item ADMIRE demonstrates strong generalization capabilities, particularly when transferring from the in-domain AndroidWorld to the out-of-domain MobileMiniWob++. It robustly yields positive improvements on both benchmarks, indicating that our method effectively integrates supervision signals without overfitting to specific task distributions.
    \vspace{-5pt}
    \item In comparison with strong baselines on AndroidWorld (Table~\ref{tab:baseline}), our method demonstrates superior performance. ADMIRE (w/Qwen2.5-VL-7B) achieves a 44.0\% success rate, surpassing GLM-4.1V-9B-Thinking and significantly outperforming the larger Qwen2.5-VL-72B, showing that 7B models can exceed closed-source or larger open-source models. Meanwhile, the smaller ADMIRE (w/Qwen2.5-VL-3B) achieves 31.0\%, competitive with 7B-scale baselines such as MobileGUI-7B, indicating that ADMIRE enables lightweight models to tackle complex GUI tasks efficiently.
    \vspace{-5pt}
\end{itemize}

\section{Analysis}
In this section, we conduct a series of experiments
to answer the following research questions (RQs):

\noindent\textbf{RQ1:} Does ADMIRE provide consistent benefits across tasks with varying levels of difficulty?

\noindent\textbf{RQ2:} Can adaptive milestones better align with the evolving policy of the agent than static milestones?

\noindent\textbf{RQ3:} How do asymmetric reward assignment (Sec ~\ref{subsec:asymmetric}) and reward decay mechanisms (Sec~\ref{subsec:grpo}) differentially affect policy optimization?

\noindent\textbf{RQ4:} Can ADMIRE generalize effectively across diverse agent tasks and reinforcement learning algorithms?

\subsection{Performance Across Varying Task Difficulties (RQ1)}
\label{sec:rq2_difficulty}

\begin{figure}[t]
    \centering
    \includegraphics[width=\linewidth]{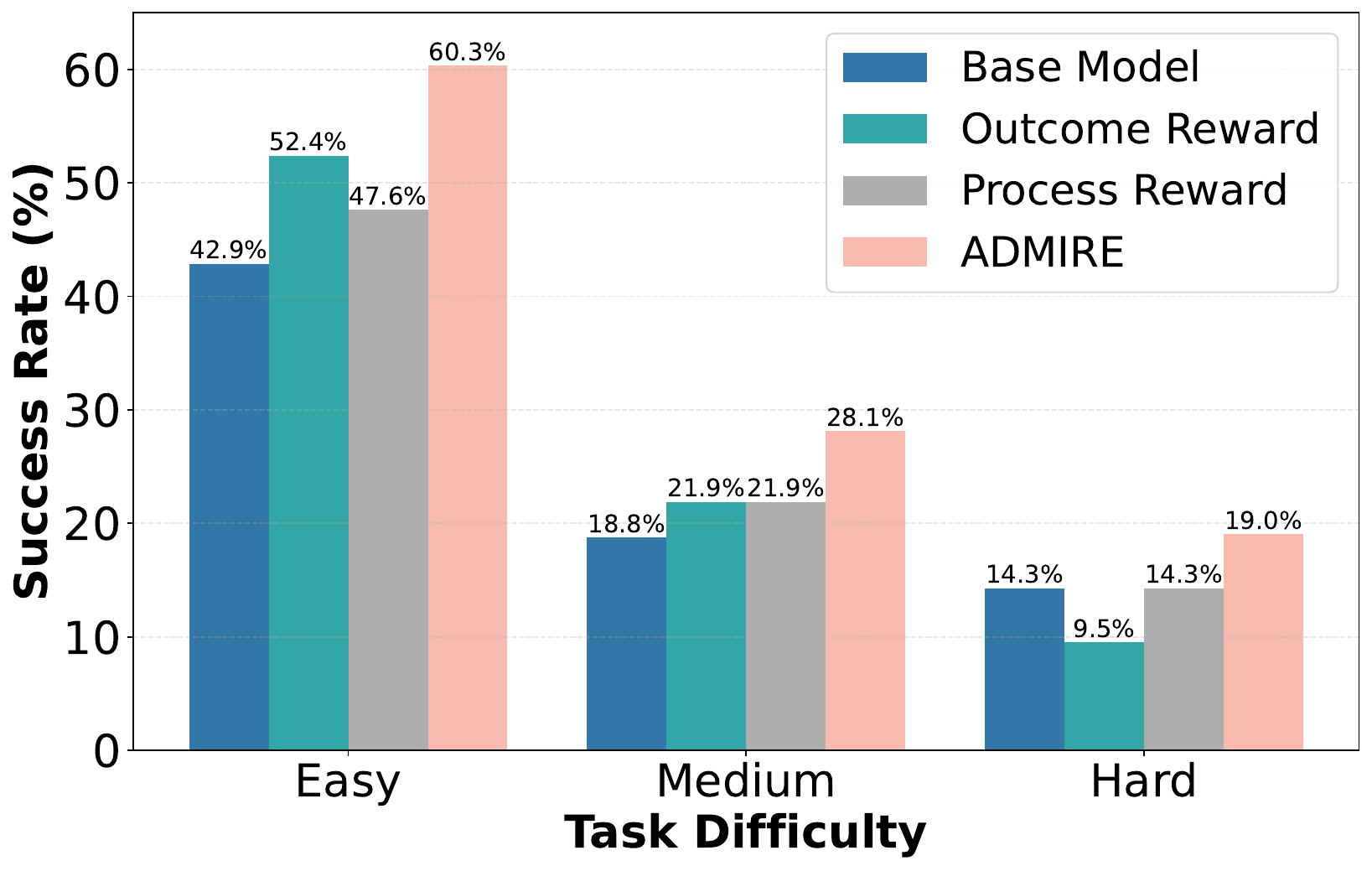}
    \vspace{-5pt}
    \caption{Comparison of success rates across different task difficulties on AndroidWorld for the base model and 7B variants trained with Outcome Reward, Process Reward, and ADMIRE. Results for the 3B model are presented in Figure~\ref{fig:difficulty_3B}.}
    \label{fig:difficulty}
    \vspace{-5pt}
\end{figure}

To evaluate ADMIRE's capabilities in long-horizon interactions, we analyze performance across different difficulty levels. As shown in Figure~\ref{fig:difficulty}, on complex, multi-step scenarios (``Hard'' tasks), outcome reward degrades performance to 9.5\% (below the 14.3\% base model), implying that sparse signals can be detrimental to pre-trained knowledge. Process reward shows no improvement over the base model. In contrast, ADMIRE achieves a 19.0\% success rate on Hard tasks, effectively mitigating credit assignment issues, demonstrating its unique effectiveness for long-horizon tasks.

Beyond complex scenarios, ADMIRE maintains superior performance across all difficulty levels. On ``Easy'' tasks, it reaches 60.3\%, outperforming Outcome (52.4\%) and Process (47.6\%) reward, and on ``Medium'' tasks, it achieves 28.1\% while the comparative methods plateau at 21.9\%. Unlike other reward mechanisms, which fluctuate or fail under high complexity, ADMIRE demonstrates robust scalability. This confirms that our framework not only excels in the complex reasoning required for long-horizon tasks but also optimizes learning efficiency for simpler interactions.

Details on task difficulty categorization are provided in the Appendix~\ref{app: difficult}.

\subsection{Effect of Adaptive Milestones (RQ2)}
\label{sec:rq3_milestone}

To evaluate the superiority of adaptive milestones over fixed ones and assess their cross-model transferability, we compare our approach against static baselines in Table~\ref{tab:milestone-compare}. Our analysis leads to the following conclusions:

\begin{itemize}[leftmargin=10pt]
    \vspace{-8pt}
    \item ADMIRE consistently outperforms static approaches across all settings. This success stems from treating milestones as adaptive entities that co-evolve with the policy. Unlike labor-intensive human annotations that might diverge from an agent’s optimal path, ADMIRE ensures intermediate rewards remain continuously aligned with emerging efficient strategies. 
    \vspace{-5pt}
    \item Milestones also demonstrate strong portability, enabling knowledge transfer from larger to smaller models. Using fixed milestones derived from the training process of a converged 7B model to train a 3B model results in decent gains (+5.9\%). This indicates that the functional states and task structures captured by the stronger model are generalizable, validating the transferability of the learned reward structures.
    \vspace{-5pt}
\end{itemize}

To further examine the impact of milestone adaptivity, we also track milestone hit counts, with detailed results provided in the Appendix~\ref{app:hit count}.

\subsection{Impact of Reward Components (RQ3)}
\label{sec:reward_analysis}

\begin{figure}[t]
    \vspace{-5pt}
    \centering
    \includegraphics[width=1\linewidth]{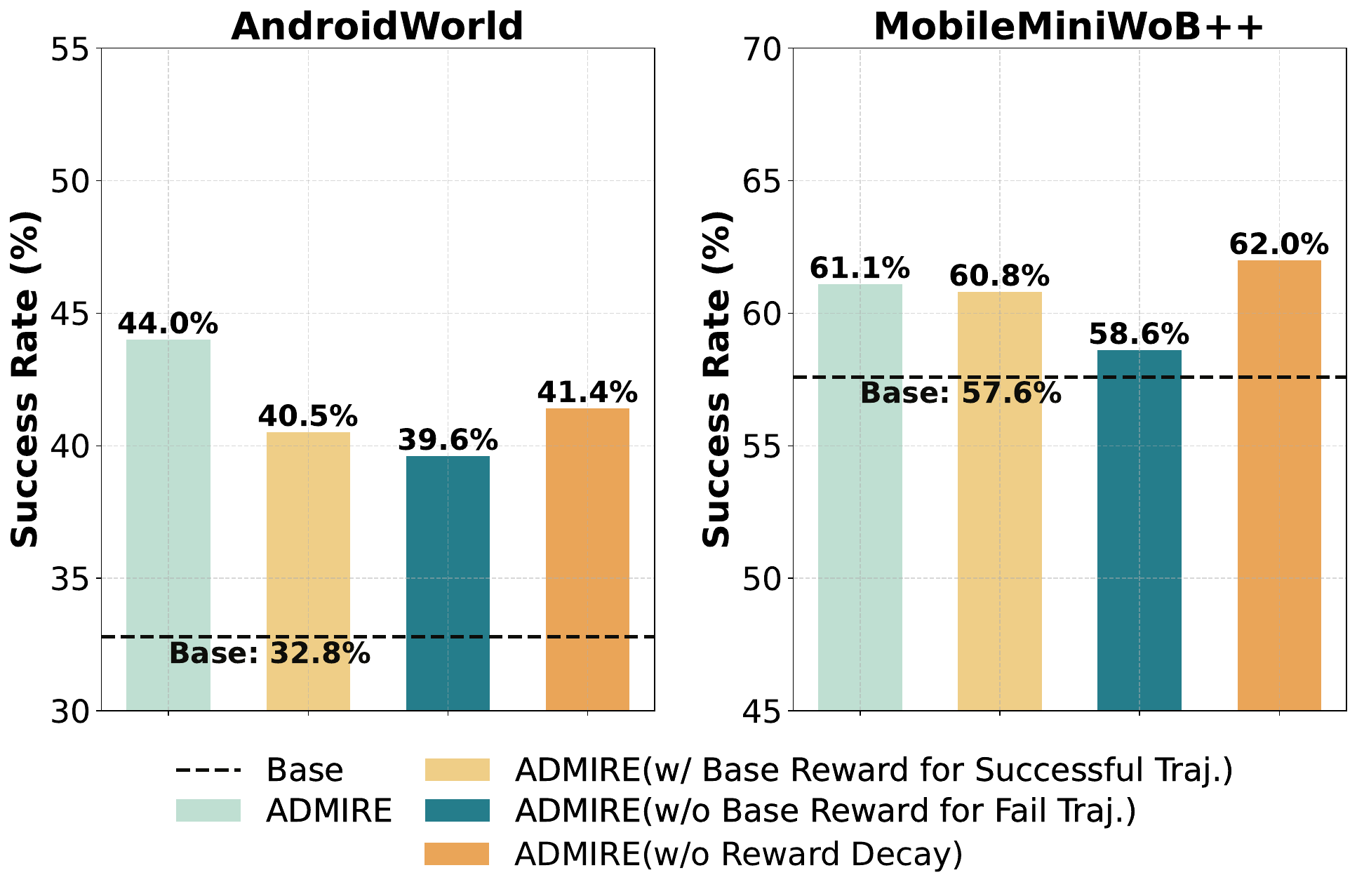}
    \vspace{-15pt}
    \caption{Ablation results comparing different ADMIRE-based reward designs and their effects on model performance. Detailed definitions of each ablation component can be found in the Appendix~\ref{app:ablation_definitions}.}
    \label{fig:ablation}
    \vspace{-10pt}
\end{figure}

To investigate the impact of different reward components, we conducted an ablation study by modifying the reward structure of ADMIRE. The performance results are presented in Figure~\ref{fig:ablation} and yield the following insights:

\begin{itemize}[leftmargin=10pt]
\vspace{-5pt}
    \item When we introduce the base reward to successful trajectories (ADMIRE w/ Base Reward for Successful Traj.), we observe a performance drop. This indicates that for successful trajectories, the binary outcome signal is already sufficient. Adding an extra dense base reward creates signal redundancy that dilutes credit assignment, distracting the optimization process rather than aiding it. This validates our strategy of assigning rewards only to critical steps in successful trajectories, ensuring a focused learning signal.
\vspace{-5pt}
    \item Removing the base progress reward from failed trajectories (ADMIRE w/o Base Reward for Fail Traj.) leads to the most significant performance degradation. This highlights that sparse milestone signals alone are insufficient for learning from failures. In complex tasks where the final goal is hard to attain, the base reward is essential for quantifying ``partial success'', acting as a necessary scaffold that guides the policy toward the goal even when the outcome score remains zero.
\vspace{-15pt}
    \item Interestingly, the variant without reward decay (ADMIRE w/o Reward Decay) does not lead to a performance decrease; it even yields a slight improvement on MobileMiniWob++ (62.0\%). This suggests that ADMIRE's adaptive milestones maintain high alignment with the policy strategy throughout the training. Consequently, strict curriculum decay is not mandatory, as the milestone guidance remains a positive signal rather than becoming noise in later training stages.
    \vspace{-5pt}
\end{itemize}

\begin{table*}[t]
\small\centering
\setlength{\tabcolsep}{7.8pt}
{
\begin{tabular}{ccccccccccc}
\toprule
\multirow{2}{*}{RL Algorithm} & \multirow{2}{*}{Reward Source} & \multicolumn{7}{c}{ALFWorld} & \multicolumn{2}{c}{WebShop} \\ 
\cmidrule(r){3-9} \cmidrule(r){10-11} 
 & & Pick & Clean & Heat & Cool & Pick2 & Look & All & Score & Succ. \\ \midrule

\multicolumn{1}{c|}{\multirow{3}{*}{GRPO}} & Outcome Reward & 83.5 & 84.5 & 75.0 & 71.4 & 52.3 & 64.3 & 75.8 & 74.7 & 50.0 \\
\multicolumn{1}{c|}{} & Process Reward & 82.4 & 81.8 & 78.6 & 72.2 & 54.5 & 66.7 & 76.6 & 75.4 & 51.7 \\
\multicolumn{1}{c|}{} & ADMIRE(Ours) & 84.8 & 85.7 & 81.8 & 73.1 & 56.3 & 63.6 & 78.1 & 81.9 & 57.0 \\ \midrule

\multicolumn{1}{c|}{\multirow{3}{*}{RLOO}} & Outcome Reward & 88.2 & 84.0 & 83.3 & \underline{82.1} & 60.7 & 73.3 & 76.1 & 80.5 & 57.0 \\
\multicolumn{1}{c|}{} & Process Reward & 84.2 & \underline{88.2} & 78.6 & 78.9 & \underline{66.7} & 66.7 & 72.7 & 80.8 & 58.6 \\
\multicolumn{1}{c|}{} & ADMIRE(Ours) & 90.0 & \textbf{95.0} & 84.0 & \textbf{88.5} & \textbf{70.4} & \textbf{76.9} & 84.4 & 80.7 & 63.3 \\ \midrule

\multicolumn{1}{c|}{\multirow{3}{*}{DAPO}} & Outcome Reward & 91.4 & 84.0 & \underline{89.5} & 79.2 & 63.1 & \underline{75.0} & 78.9 & 82.2 & 71.9 \\
\multicolumn{1}{c|}{} & Process Reward & \textbf{94.9} & \underline{88.2} & 78.6 & 80.8 & 61.1 & 72.7 & \underline{85.1} & \underline{85.5} & \underline{75.0} \\
\multicolumn{1}{c|}{} & ADMIRE(Ours) & \underline{93.8} & \textbf{95.0} & \textbf{92.3} & \textbf{88.5} & 64.7 & \textbf{76.9} & \textbf{87.5} & \textbf{85.7} & \textbf{78.1} \\

\bottomrule
\end{tabular}
}
\vspace{-5pt}
\caption{Comparison of Different Reward Settings on ALFWorld and WebShop using GRPO, RLOO and DAPO. The best result is given in \textbf{bold}, and the second-best value is \underline{underlined}.}
\label{tab:generalization}
\vspace{-8pt}
\end{table*}

\subsection{Generalizability Analysis (RQ4)}
\label{sec:generalizability}

To assess ADMIRE’s generalization ability, we extended our experiments to two diverse domains: embodied decision-making (ALFWorld~\citep{shridhar2020alfworld}) and e-commerce web agents (WebShop~\citep{yao2022webshop}) (Experimental details are provided in Appendix~\ref{app:Implementation Details for ALFWorld and WebShop}). As shown in Table~\ref{tab:generalization}, ADMIRE consistently outperforms standard reward mechanisms across these distinct environments. For instance, with the GRPO~\citep{shao2024deepseekmath}, ADMIRE achieves the highest scores on WebShop (81.9\%) and ALFWorld (78.1\%), surpassing both outcome and process reward. This suggests the core concept of adaptive milestones transcends mobile GUI tasks, capturing the essential structure inherent in various agent problems.

Furthermore, our results demonstrate that ADMIRE is compatible with different reinforcement learning algorithms. When integrated with advanced baselines like RLOO~\citep{ahmadian2024back} and DAPO~\citep{yu2025dapo}, ADMIRE exhibits even more significant performance gains. On the ALFWorld benchmark, while RLOO-ADMIRE achieves a remarkable success rate of 84.4\%, combining ADMIRE with DAPO further pushes the performance to 87.5\%. Similarly, on WebShop, the DAPO-ADMIRE combination yields the highest global success rate of 78.1\%. These results confirm that the adaptive, high-quality signals provided by ADMIRE are algorithm-agnostic, enabling different optimization objectives to exploit the improved reward structure for superior policy learning.

\subsection{Additional Analysis}
Extensive analysis yields the following conclusions:

\noindent\textbf{Hyperparameters Study} (Appendix~\ref{app:hyper}) : ADMIRE is robust to the reward weight $\lambda_0$ across a wide range of optimal values.

\noindent\textbf{Efficiency and Quality} (Appendix~\ref{app:EQ}): ADMIRE incurs negligible computational overhead and generates high-quality milestones validated by human evaluation.

\noindent\textbf{Milestone Coverage} (Appendix~\ref{app:milestone_coverage}): ADMIRE rapidly achieves milestone coverage for most tasks, providing comprehensive training supervision.

\section{Related Work}
\subsection{Mobile GUI Agents}
The rapid evolution of MLLMs~\citep{bai2025qwen2,wang2025internvl3_5,Qwen3-VL} has fundamentally transformed the landscape of autonomous mobile GUI agents. Mobile GUI Agent like~\citet{cheng2025kairos,ye2025mobile,li2025mobileuse,li2025coloragent,wu2025verios, ma-etal-2024-coco} have demonstrated remarkable capabilities in executing human-like gestures. Despite these impressive perceptual capabilities, agents continue to struggle with complex, long-horizon tasks~\citep{song2025colorbench, ma-etal-2025-caution} in real-world environments. Consequently, the research focuses on post-training paradigms to enhance robustness and generalization, and shift Supervised Fine-Tuning (SFT) towards Reinforcement Learning(RL). Approaches like ~\citet{gu2025mobile,xumobilerl,lu2025arpo,lu2025ui} employ online or offline RL to encourage active exploration and self-correction. However, mobile agent tasks typically provide only sparse binary outcomes, making it difficult for the agent to learn efficient policies without dense, high-quality feedback. This challenge motivates our work, which seeks to establish a more effective RL training pipeline through adaptive, verifiable reward shaping.

\subsection{Process Reward Supervision}
To mitigate the sparsity inherent in outcome-based rewards, recent research has increasingly turned toward Process Reward Models (PRMs) \citep{lightman2023let,zheng2025survey}. While pioneering in mathematical reasoning tasks \citep{wang2024math,zhu2025retrieval,zheng2025cold,zhang2025lessons,zou2025reasonflux}, PRMs evaluate the validity of intermediate steps to provide granular supervision. In the realm of GUI navigation, GUI-Shepherd \cite{chen2025gui} offers dense, step-by-step feedback, while GUI-PRA \cite{xiong2025gui} generates process rewards by intelligently analyzing historical context and UI state transitions. However, extending process supervision to open-ended mobile environments introduces unique challenges, most notably reward hacking~\citep{taylor2025school}, where agents exploit verifier biases to maximize rewards without meaningful progress toward the goal. To combat this, our work establishes a verifiable and objective milestone-based reward framework.

\section{Conclusion}
In this work, we address the temporal credit assignment challenge in Mobile GUI Agents by proposing the \textbf{Ad}aptive \textbf{Mi}lestone \textbf{Re}ward (ADMIRE) mechanism, which bridges the gap between sparse outcome signals and noisy process supervision through verifiable, adaptive milestones. By synergizing dynamic milestone generation with an asymmetric credit assignment strategy, ADMIRE effectively denoises successful trajectories while providing essential scaffolding for failed attempts, ensuring dense yet high-fidelity feedback. Extensive experiments on AndroidWorld demonstrate that our approach significantly enhances learning efficiency, enabling 7B parameter models to outperform 72B baselines. Furthermore, ADMIRE exhibits exceptional robustness and generalization across cross-domain benchmarks like ALFWorld and WebShop, establishing a scalable, algorithm-agnostic paradigm for training agents to master complex, long-horizon tasks.

\section*{Limitations}
ADMIRE has two main limitations. First, the effectiveness of ADMIRE is inherently bound by the reasoning capabilities of the VLM employed for milestone generation; although extracting high-level sub-goals is a relatively straightforward summarization task compared to precise execution, the system remains somewhat dependent on the generator's quality. Furthermore, our current framework utilizes these verifiable milestones exclusively for auxiliary reward assignment, leaving their potential as a proxy outcome signal unexplored for open-ended environments that lack rule-based success detection mechanisms.



\bibliography{custom}

\appendix

\input{appendix}

\end{document}

%% file: appendix.tex
\begin{algorithm*}[t]
\caption{ADMIRE: Adaptive Milestone Reward Mechanism with GRPO}
\label{alg:admire}
\begin{algorithmic}[1]
\Require Dataset $\mathcal{G}$, Policy $\pi_\theta$,LLM $\Phi$ , BERT $\psi$
\Require Hyperparameters: $\lambda_0$ (reward weight), $\gamma$ (decay), $\eta$ (format weight), $\delta$ (similarity threshold)
\State Initialize global milestone memory $\mathcal{M} \leftarrow \emptyset$ 
\State Initialize policy parameters $\theta$

\For{epoch $\mathcal{E} = 1, 2, \dots, E_{\text{max}}$}
    \State Update curriculum coefficient $\lambda(t) \leftarrow \lambda_0 \cdot \gamma^{\mathcal{E}}$
    \For{sampled instruction batch $G \sim \mathcal{G}$}
        \State $\mathcal{B} \leftarrow \text{Rollout}(\pi_{\theta}, G, B)$ \Comment{Generate $B$ trajectories}
        \State $\mathcal{B}_G^+ \leftarrow \{ \tau \in \mathcal{B} \mid \mathcal{O}(\tau) = 1 \}$ \Comment{Filter successful trajectories}
        
        \Statex \textbf{// Phase 1: Adaptive Milestone Generation \& Evolution}
        \If{$\mathcal{B}_G^+ \neq \emptyset$}
            \State Select superior trajectory $\tau_{\text{new}}$ from $\mathcal{B}_G^+$
            \If{$G \notin \mathcal{M}$ or $\mathcal{M}_G$ is empty} \Comment{Case: Initialization}
                \State $\mathcal{M}_G^{(0)} \leftarrow \Phi(\tau_{\text{new}}, G, \mathbf{P}_{\text{init}})$
            \Else \Comment{Case: Refinement}
                \State $\mathcal{M}_G^{(i+1)} \leftarrow \Phi(\tau_{\text{new}}, \mathcal{M}_G^{(i)}, G, \mathbf{P}_{\text{update}})$
            \EndIf
        \EndIf
        
        \Statex \textbf{// Phase 2: Reward Calculation via Asymmetric Credit Assignment}
        \State Initialize Advantage buffer $\mathcal{A} \leftarrow []$
        \For{trajectory $\tau \in \mathcal{B}$}
            \State Let $K \leftarrow |\mathcal{M}_G|$, match count $k \leftarrow 0$, pointer $p \leftarrow 1$
            \For{step $t = 1, \dots, T$ in $\tau$}
                \State Calculate $s(a_t, m_p) \leftarrow \frac{\psi(a_t) \cdot \psi(m_p)}{\| \psi(a_t) \| \| \psi(m_p) \|}$ \Comment{Semantic Similarity}
                \State $r^{\text{mil}}_t \leftarrow 0$
                \If{$s(a_t, m_p) > \delta$} \Comment{Hit Milestone}
                    \State $r^{\text{mil}}_t \leftarrow s(a_t, m_p)$
                    \State $k \leftarrow k + 1$, $p \leftarrow \min(p+1, K)$
                    \State Mark step $t$ as milestone hit ($t \in \mathcal{T}_{\text{mil}}$)
                \EndIf
                
                \Statex \qquad\qquad \textit{// Apply Asymmetric Strategy}
                \If{$\mathcal{O}(\tau) = 1$} \Comment{Case 1: Denoising Positive Samples}
                    \State $\mathcal{R}_{\text{mil}}(t) \leftarrow (t \in \mathcal{T}_{\text{mil}}) ~?~ r^{\text{mil}}_t : 0$
                \Else \Comment{Case 2: Scaffolding Negative Samples}
                    \State $\mathcal{R}_{\text{mil}}(t) \leftarrow \frac{k}{K} + \left( (t \in \mathcal{T}_{\text{mil}}) ~?~ 0.5 \cdot r^{\text{mil}}_t : 0 \right)$
                \EndIf
                
                \State $\mathcal{R}_{\text{total}}(t) \leftarrow \mathcal{R}_{\text{outcome}}(t) + \eta \cdot \mathcal{R}_{\text{format}}(t) + \lambda(t) \cdot \mathcal{R}_{\text{mil}}(t)$
            \EndFor
            \State Compute Advantage $\hat{A}$ using $\mathcal{R}_{\text{total}}$
            \State $\mathcal{A}.\text{append}(\hat{A})$
        \EndFor
        
        \Statex \textbf{// Phase 3: Optimization}
        \State Update $\theta$ via GRPO loss $\mathcal{J}(\theta)$ using $\mathcal{B}$ and $\mathcal{A}$ (Eq.~\ref{eq:grpo_loss})
    \EndFor
\EndFor
\end{algorithmic}
\end{algorithm*}

\section{Experiments Setting}
\subsection{Implementation Details}
\label{app:Implementation Details}

\subsubsection{Infrastructure and Hardware}
To support online training, we launch Android emulator instances on remote machines, exposing dedicated ports to establish interactions via IP addresses. We train online in parallel with 32 emulators simultaneously to maximize data throughput. The training process is accelerated using 8 NVIDIA A800 (80GB) GPUs.

\subsubsection{Training Configuration}
In each online RL iteration, the system samples 4 distinct tasks and collects 8 full trajectories via parallel rollouts across 32 remote emulators. To manage complexity and computational overhead, the maximum trajectory length is capped at 20 steps. Furthermore, we constrain the maximum input prompt and generated response lengths to 6500 and 512 tokens, respectively.

Regarding the reward configuration, we set the matching threshold $\delta=0.75$ based on the analysis in Section~\ref{exp:threshold}. We configure $\zeta=0.5,\eta=0.5$ and the initial weight parameter $\lambda_0=0.3$ with a decay factor of $\gamma=0.99$. Additionally, the weight assigned to the process reward is fixed at 0.3.

For the optimization process, we employ a constant learning rate of $1\times10^{-6}$ and a PPO clip ratio of 0.2. The model is updated for 2 epochs per iteration with a mini-batch size of 128. Finally, we apply advantage normalization to stabilize training but exclude both the KL divergence loss and the entropy regularization coefficient.

\subsubsection{Model Architecture and Auxiliary Components}
The GUI agent accepts inputs consisting of a sequence of past action descriptions and visual information from the current screenshot. To facilitate decision-making, the agent is designed to first externalize its reasoning process before invoking a predefined mobile function interface, which specifies the supported action types and required parameters. To mitigate the limited instruction-following capabilities of the base models and ensure stability, we conduct a warm-up phase for both Qwen2.5-VL variants prior to online RL, following the protocol described by~\citet{li2025coloragent}. The detailed prompts used for the GUI agent are illustrated in Figures~\ref{fig:prompt_gui_part1}--\ref{fig:prompt_gui_part3}.

In addition to the primary policy model, we incorporate auxiliary models to support the training loop. We employ GPT-4o~\citep{achiam2023gpt} to initialize and refine task milestones, as well as to serve as an LLM-as-Judge for providing process reward signals. The prompts governing milestone initialization and refinement are presented in Figure~\ref{fig:Prompt initialize} and Figure~\ref{fig:Prompt Refine}, respectively. Note that the instructions issued to human annotators are identical to these prompts. Furthermore, we utilize a Sentence-BERT~\citep{reimers2019sentence} model to generate embeddings, which are used to compute the semantic similarity between milestones and agent actions.

\subsubsection{Implementation Details for ALFWorld and WebShop}
\label{app:Implementation Details for ALFWorld and WebShop}
For the generalization experiments, we conduct online reinforcement learning using a framework built upon verl-agent~\citep{feng2025group}. We employ Qwen2.5-1.5B-Instruct as the base model, and the training is executed on 4 NVIDIA A800 (80GB) GPUs.

The training is conducted with a constant learning rate of $1 \times 10^{-6}$ and a KL divergence coefficient of $\beta=0.01$ to maintain training stability. For generation and exploration, we set the group size to $G=8$. 

\subsection{Details on Benchmarks and Baselines}
\label{app:benchmark baseline}
\subsubsection{Benchmarks}
In this section, we provide a more detailed overview of the benchmarks.
\begin{itemize}
    \item \textbf{AndroidWorld}~\citep{rawles2024androidworld} is a dynamic benchmark designed for GUI agents within the Android ecosystem, featuring 116 multi-step tasks across 20 real-world applications. Leveraging Android Studio to emulate a Pixel 6 (Android 13, API 33), it provides a realistic online environment where success is determined via programmatic checks. To ensure reproducibility, AndroidWorld utilizes task templates where specific instances are controlled by random seeds. Tasks are further categorized into easy, medium, and hard difficulty levels, facilitating a granular evaluation of agent capabilities.
    \item \textbf{MobileMiniWoB++ }is a mobile-centric web benchmark adapted by ~\citet{rawles2024androidworld} from the original MiniWoB++ ~\citep{liu2018reinforcement}. It consists of 92 tasks integrated within a single simulated application, focusing on localized interactions rather than multi-page navigation. Typical of web-based benchmarks, these tasks feature a high density of UI elements, posing a significant challenge to an agent’s ability to accurately localize elements.
    \item \textbf{ALFWorld}~\citep{shridhar2020alfworld} is an embodied simulator designed to assess the multi-step decision-making and language reasoning capabilities of agents within interactive household environments. The benchmark requires agents to achieve specific text-based goals through multi-turn interactions across 3,827 task instances, which are categorized into six distinct household activities: picking and placing objects, examining items under light, cleaning, heating, cooling, and coordinating multiple objects. Evaluation is centered on the agent's ability to translate high-level natural language instructions into a sequence of grounded actions, with success determined by the completion of the state-based objectives defined within these diverse activity categories.
    \item \textbf{WebShop}~\citep{yao2022webshop} is an interactive e-commerce benchmark designed to evaluate an agent's language grounding and decision-making across 1.18 million real-world products and 12,087 crowd-sourced instructions. To succeed, an agent must navigate complex web interfaces, perform query reformulation, and execute precise actions to find and customize items matching specific user requirements. Evaluation is conducted through an automated reward function that programmatically measures the attribute-level alignment between the purchased product and the initial instruction, providing an objective metric for multi-step task completion in a high-density UI environment.
\end{itemize}

\subsubsection{Baselines}
In this section, we provide a more detailed overview of the baselines.
\paragraph{Reward Baseline}
\begin{itemize}
    \item \textbf{Outcome reward} is a binary reward mechanism based on predefined rules. It is granted only at the termination of a task, where $r=1$ signifies success and $r=0$ signifies failure.
    \item \textbf{Process Reward} is a model-based reward mechanism that provides a scalar score for every step within a task trajectory. In our implementation, we employ an LLM-as-a-Judge framework, utilizing GPT-4o~\citep{achiam2023gpt} as the evaluator. The specific prompt configuration used for the evaluation model is detailed in Figure ~\ref{fig:Prompt PRM}.
\end{itemize}

\paragraph{Model Baseline}
\begin{itemize}
    \item \textbf{GPT-4o-2024-11-20}~\citep{achiam2023gpt}is OpenAI’s flagship multimodal model designed for real-time, native interaction across text, audio, and images, offering human-like response latency and state-of-the-art performance in reasoning and creative tasks.
    \item \textbf{Claude-Sonnet-4-20250514-thinking}~\citep{claude4} is a hybrid reasoning model from Anthropic that utilizes an extended thinking mechanism to perform deep, step-by-step internal reasoning before providing outputs, significantly enhancing its performance in complex coding, multi-step agentic workflows, and nuanced problem-solving.
    \item \textbf{InfiGUIAgent}~\citep{liu2025infiguiagent}is an MLLM-based agent trained via a two-stage supervised fine-tuning pipeline that first establishes foundational GUI understanding and grounding skills, then integrates hierarchical and expectation-reflection reasoning through synthesized data to achieve native multi-step decision-making capabilities.
    \item \textbf{OS-Genesis}~\citep{sun2025genesis} is a GUI agent trained via a novel retrospective synthesis pipeline that reverses the traditional data collection process, allowing the model to derive high-quality tasks from autonomous environment exploration and ensuring superior performance through a dedicated trajectory reward model.
    \item \textbf{GUI-PRA}~\citep{xiong2025gui} is an advanced evaluator designed to provide fine-grained process rewards by integrating a dynamic memory mechanism—comprising relevance-based retrieval and progressive summarization—with an adaptive UI perception system that actively reasons about state changes to gather grounded visual evidence.  
    \item \textbf{Aguvis}~\citep{xu2024aguvis} is a unified, vision-based framework that enables generalizable interface understanding by operating directly on screen images and utilizing a standardized plugin-based action space, while incorporating an explicit inner monologue during training to foster sophisticated, human-like reasoning patterns across diverse GUI environments.
    \item \textbf{GUI-Critic-R1}~\citep{wanyan2025look} is a specialized model designed for efficient GUI pre-criticism that employs Suggestion-aware Group Relative Policy Optimization (S-GRPO) and an innovative suggestion reward to refine its reasoning and provide reliable guidance for correcting erroneous operations. 
    \item \textbf{MobileGUI-7B}~\citep{shi2025mobilegui} is a GUI agent trained via a novel reinforcement learning framework that leverages an interactive virtual machine environment for continuous online learning, utilizing a synthetic task generation pipeline and an adapted GRPO algorithm with trajectory-aware rewards to balance task success and execution efficiency.
    \item \textbf{UI-TARS-1.5-7B}~\citep{qin2025ui} is a state-of-the-art GUI agent that combines enhanced context-aware perception and unified action modeling with "System-2" reasoning patterns—such as task decomposition and milestone recognition—and utilizes an iterative training pipeline with reflective online traces to continuously learn from mistakes across diverse environments.
    \item \textbf{Qwen2.5-VL-72B}~\citep{bai2025qwen2} is Alibaba's flagship vision-language model, featuring 72 billion parameters and state-of-the-art capabilities in complex visual understanding, long-video analysis, and high-precision document parsing.
    \item \textbf{GUI-Shepherd}~\citep{chen2025gui} is a process reward model trained on 52k high-quality interactions that provides step-by-step feedback and rationales to guide agents in complex GUI tasks. 
    \item \textbf{GLM-4.1V-9B-Thinking}~\citep{hong2025glm} is a vision-language model optimized for advanced multimodal reasoning through a unified framework that combines knowledge-intensive pre-training with Reinforcement Learning with Curriculum Sampling (RLCS).
\end{itemize}

\section{Supplementary Experiments}
\subsection{Performance Across Varying Task Difficulties}
\label{app: difficult}
The task difficulty levels in AndroidWorld~\citep{rawles2024androidworld} are classified into Easy, Medium, and Hard categories based on empirical evaluations provided by human annotators. After performing each assigned task to establish a performance baseline, annotators assigned a difficulty level primarily governed by the interaction trajectory length and the scope of the workflow. 

\begin{figure}[t]
    \centering
    \includegraphics[width=\linewidth]{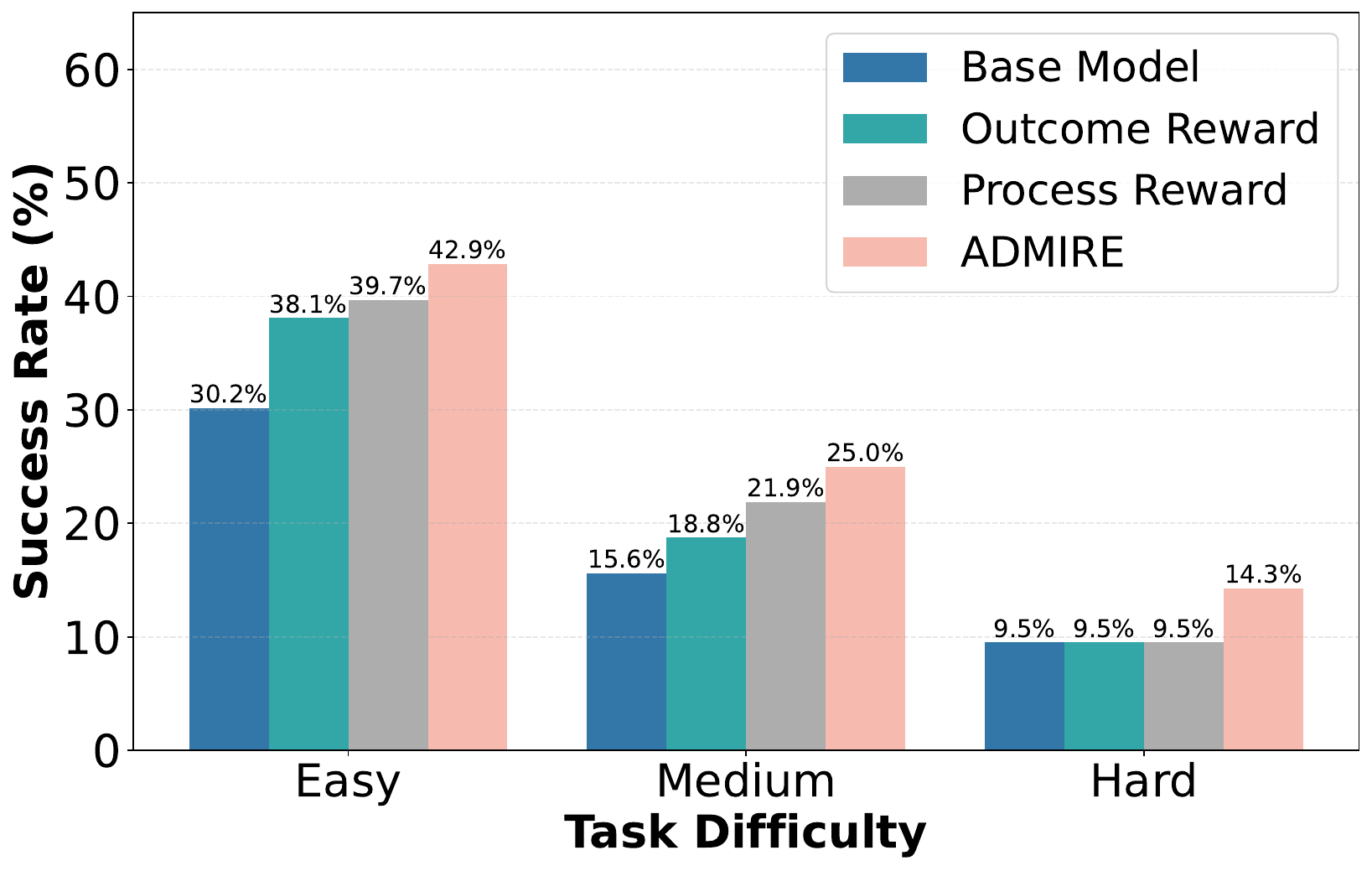}
    \vspace{-5pt}
    \caption{Comparison of success rates across different task difficulties on AndroidWorld for the base model and models trained with outcome, process reward, and ADMIRE.}
    \label{fig:difficulty_3B}
    \vspace{-5pt}
\end{figure}

To better illustrate the effectiveness of ADMIRE on long-horizon tasks, we also report experiments conducted with the 3B model, as shown in Figure~\ref{fig:difficulty_3B}.
\subsection{Dynamic and Static Milestones Hit Count Analysis}
\label{app:hit count}
To provide a microscopic view of how milestone adaptivity influences training dynamics, we conducted a case study on the \texttt{MarkorDeleteNote} task, tracking the hit counts of intermediate milestones throughout the training process.

In the Static setting, milestone hits are heavily skewed towards the initial stages (the first two milestones), with a sharp decline in activation for the subsequent milestones. This phenomenon indicates a critical policy-reward misalignment: as the agent's policy evolves, its exploration trajectory inevitably drifts away from the fixed path defined by static milestones. Once the agent deviates from this pre-defined rigid path, it fails to trigger subsequent static checkpoints, effectively causing the dense reward signal to vanish halfway through the task.

In contrast, Dynamic milestones exhibit a consistently high activation frequency across all four stages of the task, even in the late training phases. This confirms that our co-evolutionary mechanism successfully bridges the gap between the reward structure and the agent's behavior. By updating milestones to match the agent's efficient trajectories, ADMIRE ensures that supervision remains dense and continuous throughout the entire long-horizon episode, preventing the "signal attenuation" problem that plagues static reward shaping.

\begin{figure}[ht]
    \centering
    \includegraphics[width=1\linewidth]{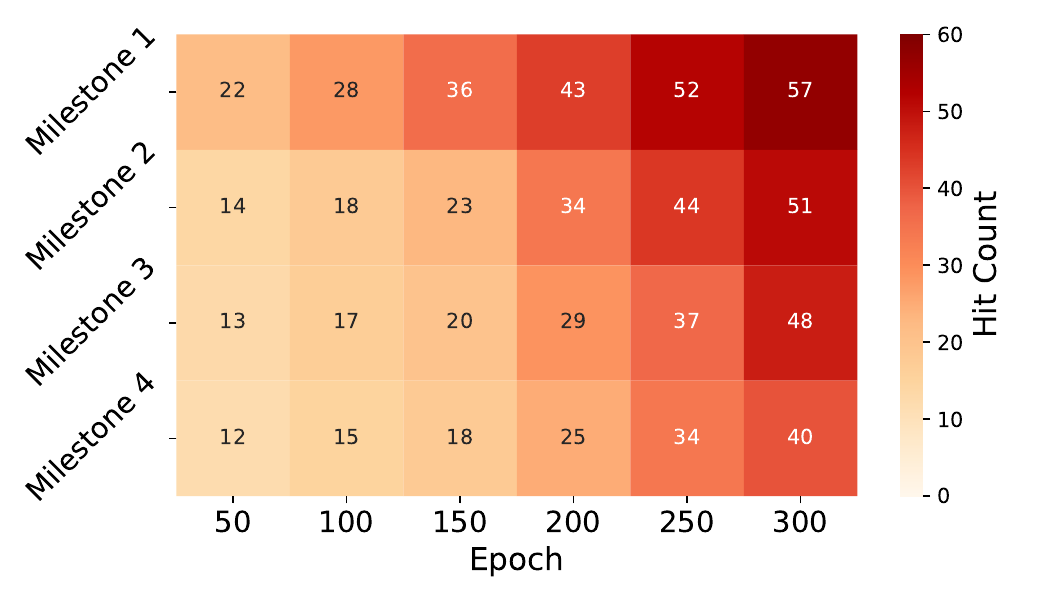}
    \caption{Dynamic milestone hit count during training.}
    \label{fig:hitcount}
\end{figure}

\begin{figure}[ht]
    \centering
    \includegraphics[width=1\linewidth]{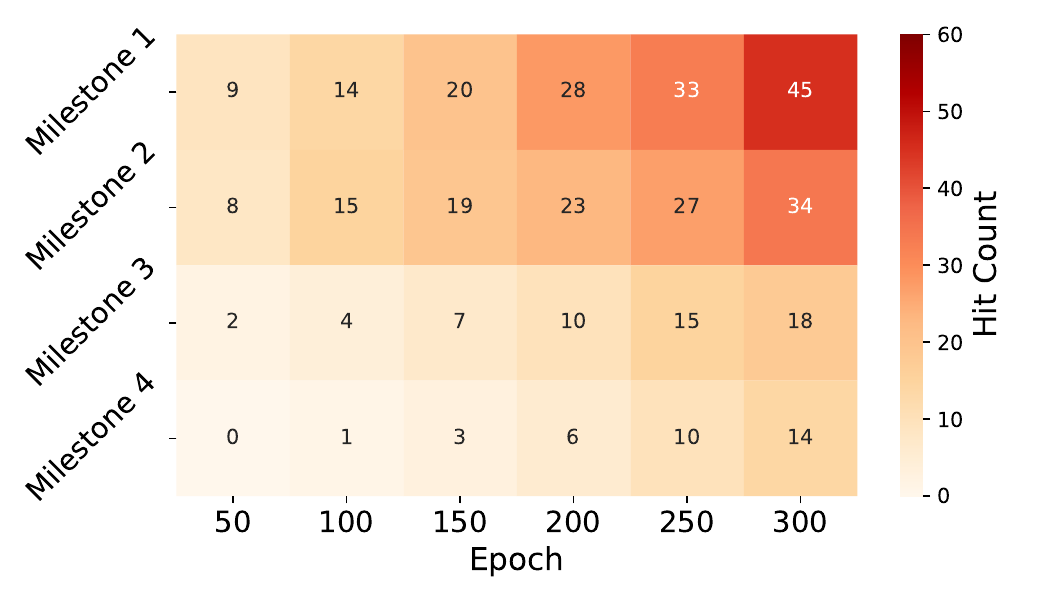}
    \caption{Static milestone hit count during training.}
    \label{fig:hitcount_static}
\end{figure}

\subsection{Definitions of Ablation Variants}
\label{app:ablation_definitions}

In Section~\ref{sec:reward_analysis}, we compare the full ADMIRE framework against three specific variants to validate our reward design. The detailed configurations of these variants are defined as follows:

\begin{itemize}
    \item \textbf{ADMIRE (w/ Base Reward for Successful Traj.):} 
    In the standard ADMIRE design, successful trajectories ($\mathcal{O}(\tau)=1$) receive only sparse milestone reward to reduce noise. This variant alters that design by introducing the continuous base progress reward ($k/K$) even for successful episodes. This allows us to test whether dense supervision is beneficial when a strong outcome signal is already present.
    
    \item \textbf{ADMIRE (w/o Base Reward for Fail Traj.):} 
    Standard ADMIRE employs a "scaffolding" mechanism where failed trajectories ($\mathcal{O}(\tau)=0$) receive a base reward ($k/K$) to quantify partial success. This variant removes that component, forcing the model to learn from failed trajectories using the sparse milestone hits solely without the continuous progress indicator.
    
    \item \textbf{ADMIRE (w/o Reward Decay):} 
    The standard design includes a time-dependent curriculum coefficient $\lambda(t)$ that decays the weight of the milestone reward over epochs. This variant removes the decay mechanism (setting $\gamma=1$), keeping the weight of the milestone reward constant throughout the entire training process to evaluate the necessity of shifting focus to outcome reward.
\end{itemize}

\subsection{Hyperparameter Study}
\label{app:hyper}
\subsubsection{The Effect of Reward Weight \texorpdfstring{$\lambda_0$}{lambda}}

\begin{figure}[ht]
    \centering
    \includegraphics[width=1\linewidth]{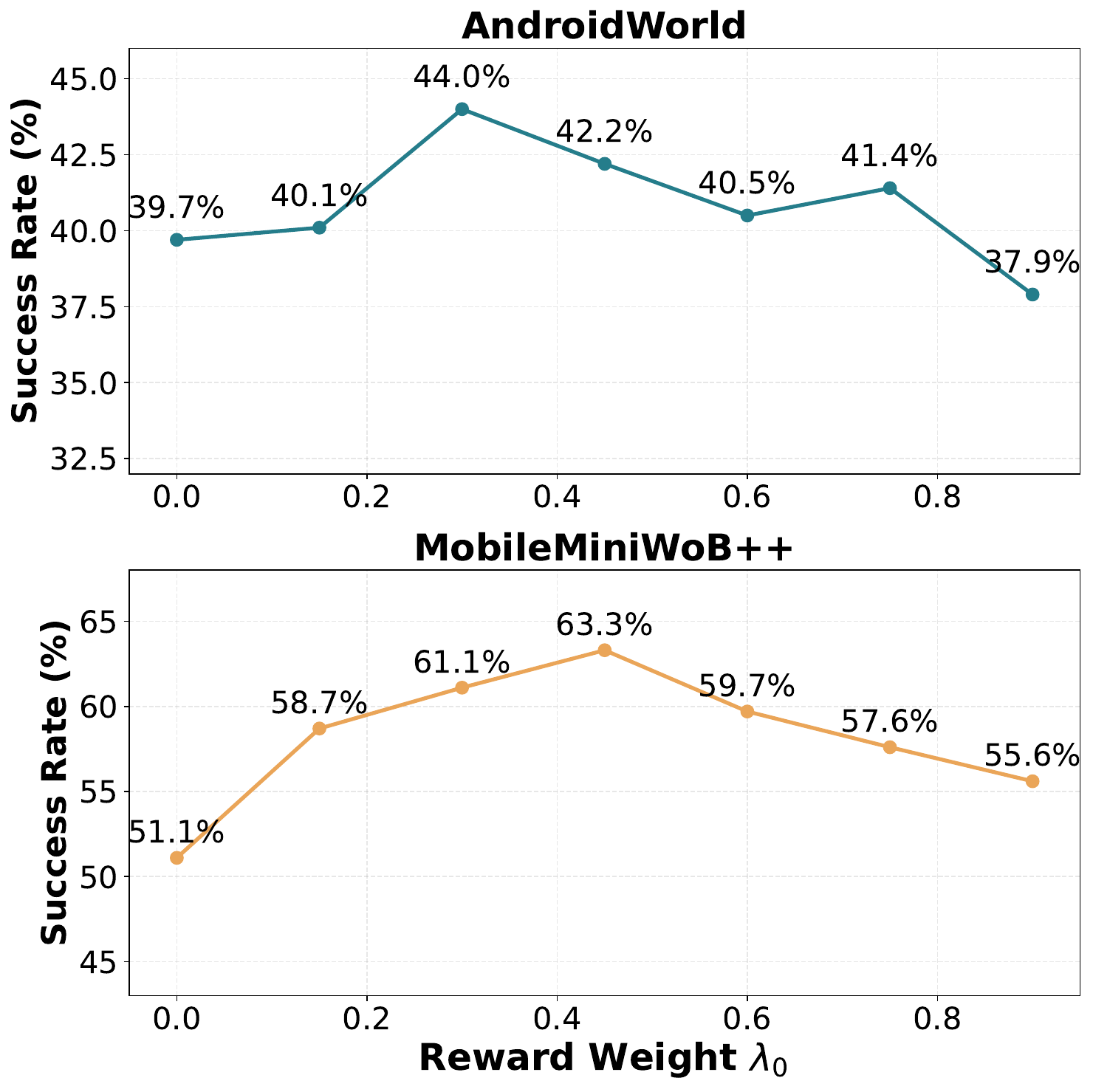}
    \caption{Impact of ADMIRE on RL training across varying reward weights $\lambda_0$.}
    \label{fig:hyper lambda}
\end{figure}

We further investigate the sensitivity of ADMIRE to the initial milestone reward weight $\lambda_0$, which balances the dense supervision against the sparse outcome signal. As illustrated in Figure~\ref{fig:hyper lambda}, the performance exhibits an inverted U-shaped trend on both benchmarks. 

At the lower end ($\lambda_0=0$), the model relies solely on sparse outcome rewards, resulting in suboptimal performance (e.g., 39.7\% on AndroidWorld). Increasing $\lambda_0$ introduces necessary guidance, boosting the success rate significantly, with peaks observed at 0.3 for AndroidWorld (44.0\%) and 0.45 for MobileMiniWob++ (63.3\%). Conversely, excessively high weights (e.g., $\lambda_0=0.9$) lead to a performance decline, likely because the intermediate rewards begin to overshadow the final task objective. 
Crucially, however, the results demonstrate that ADMIRE is highly robust to hyperparameter variations. The success rates remain consistently high across a broad range of values (e.g., $\lambda_0 \in [0.3, 0.75]$ on AndroidWorld), indicating that our method does not require extensive fine-tuning to achieve superior results.

\subsubsection{The Effect of Similarity Threshold \texorpdfstring{$\delta$}{delta}}
\label{exp:threshold}
To validate the reliability and efficiency of our embedding-based milestone matching mechanism, we constructed a validation dataset consisting of 500 milestone-action description pairs. These pairs were manually annotated to establish ground truth. The distribution of positive (matched) and negative (unmatched) samples is detailed in Table~\ref{tab:matching_data}.

\begin{table}[ht]
\centering
\begin{tabular}{lcc}
\toprule
Category & Definition & Count  \\ 
\midrule
Positive Samples & Matched & 254\\
Negative Samples & Unmatched & 246  \\ 
\midrule
\textbf{Total} & - & \textbf{500}  \\ 
\bottomrule
\end{tabular}
\caption{Distribution of the manually annotated validation dataset used for evaluating matching accuracy.}
\label{tab:matching_data}
\end{table}

\begin{figure}[ht]
    \centering
    \includegraphics[width=1\linewidth]{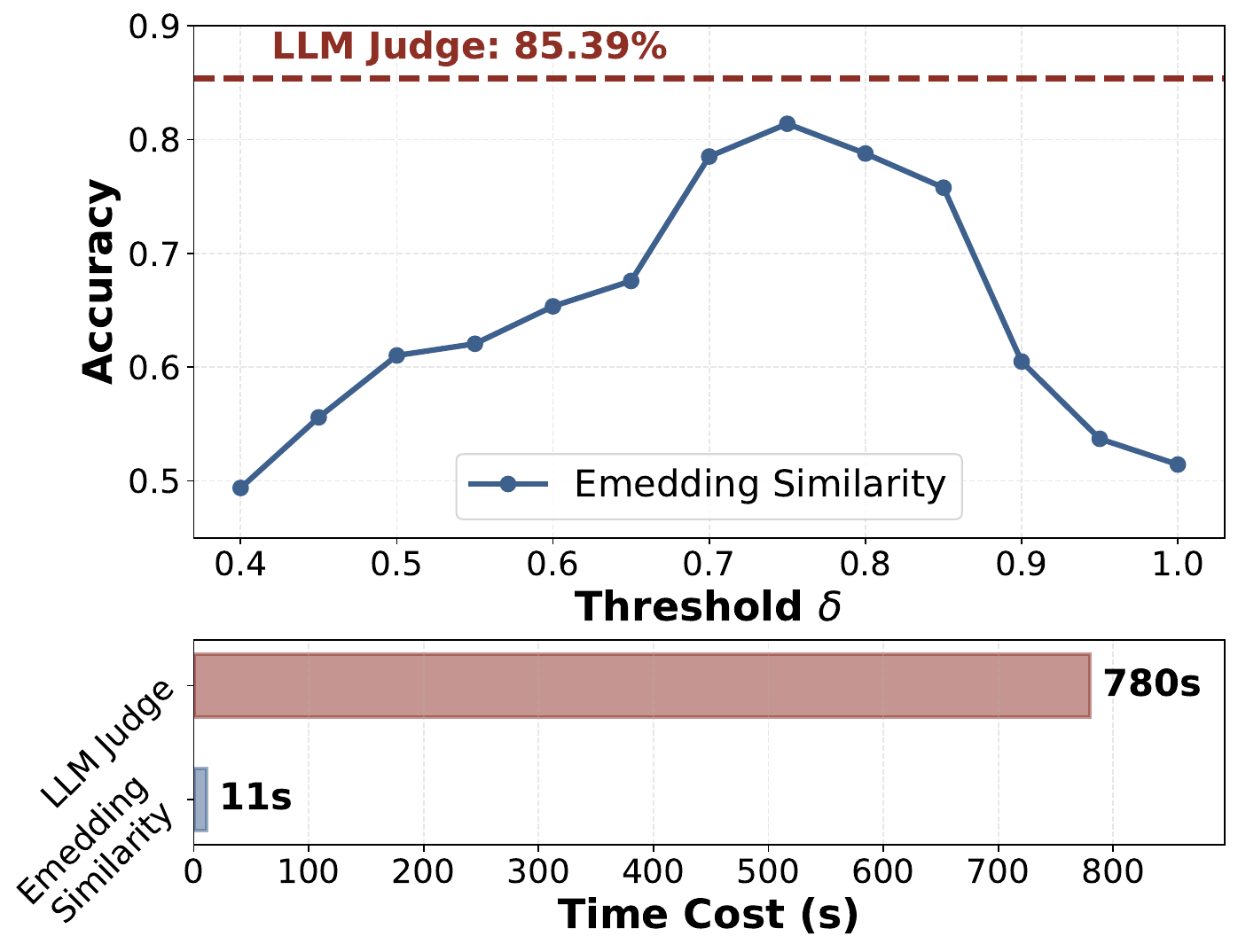}
    \caption{Effect of Similarity Threshold $\delta$ on the Accuracy of Milestone–Action Description Matching, Compared with LLM-Based Matching Methods}
    \label{fig:hyper delta}
\end{figure}

We compared our BERT-based embedding similarity method with a strong LLM-based judge (GPT-4o as the oracle), using the prompt shown in Figure~\ref{fig:Prompt match}. Figure~\ref{fig:hyper delta} presents the accuracy trends and time costs. Our analysis reveals two key observations:

As shown in the top chart of Figure~\ref{fig:hyper delta}, the matching accuracy is sensitive to the threshold $\delta$. The performance follows an inverted U-shape, peaking at $\delta = 0.75$. At this optimal threshold, the embedding method achieves an accuracy of 81.3\%. Although this is slightly lower than the LLM Judge's performance (85.39\%), the gap is marginal, demonstrating that a simple semantic similarity check is sufficiently robust for distinguishing correct actions from incorrect ones in most cases. Crucially, this reliability is further reinforced in practice by our sequential causality mechanism: by maintaining a pointer $p_t$ to track the next uncompleted milestone, we effectively prevent out-of-order matching, thereby significantly enhancing the precision of the verification process.

The bottom chart of Figure~\ref{fig:hyper delta} reveals the decisive advantage of the embedding approach: computational efficiency. Processing the validation set took the LLM Judge approximately 780 seconds, whereas the embedding similarity method completed the task in just 11 seconds—a speedup of over 70$\times$.
Given that reward calculation occurs at every step of the reinforcement learning process, the high latency of an LLM Judge is prohibitive. Therefore, the embedding-based method with $\delta=0.75$ offers the optimal trade-off, providing high-quality supervision signals with negligible computational overhead.

It is worth noting that although the overall curve indicates that matching accuracy is sensitive to the threshold $\delta$, determining the optimal value for different environments remains a straightforward process. We can readily identify the optimal threshold by collecting a small batch of task-specific sentence matching pairs and conducting a simple, rapid search experiment. This offline calibration strategy effectively circumvents the need for parameter searching during the computationally expensive online reinforcement learning phase, thereby ensuring both adaptability and training efficiency.

\subsection{Efficiency and Quality Analysis}
\label{app:EQ}
\subsubsection{Time Consumption Comparison}

\begin{figure*}[ht]
    \centering
    \includegraphics[width=\textwidth]{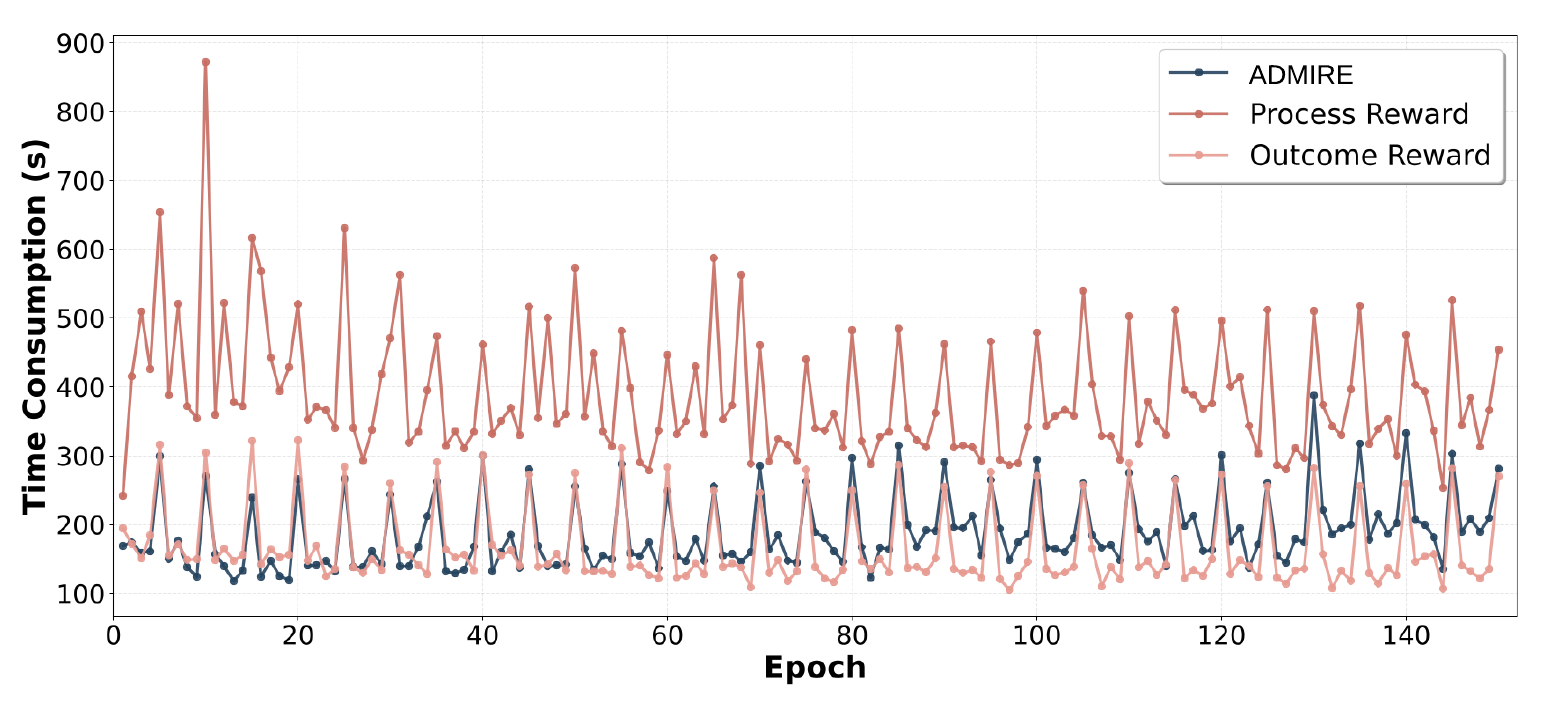}
    \caption{Per-step time consumption of different reward mechanisms during training.}
    \label{fig:time consumption}
\end{figure*}

\begin{table*}[ht]
\centering
\resizebox{\textwidth}{!}{
\begin{tabular}{cll}
\toprule
\textbf{Score} & \textbf{Rating} & \textbf{Description \& Criteria} \\ 
\midrule
\textbf{5} & \textbf{Perfect} & 
\begin{tabular}[c]{@{}l@{}}
The milestones are completely accurate, logically ordered.\\ 
Perfectly abstracts the key functional steps without redundancy.\\ 
Matches the "Gold Standard" of human expert reasoning.
\end{tabular} \\ 
\midrule
\textbf{4} & \textbf{Good} & 
\begin{tabular}[c]{@{}l@{}}
Factually correct and helpful for task completion.\\ 
May contain minor phrasing issues or slightly sub-optimal granularity \\ 
(e.g., one milestone split into two), but the logic is sound.
\end{tabular} \\ 
\midrule
\textbf{3} & \textbf{Acceptable} & 
\begin{tabular}[c]{@{}l@{}}
Usable, but contains noise. \\ 
Includes some redundant steps or misses a minor action, \\ 
but generally guides the agent in the correct direction.
\end{tabular} \\ 
\midrule
\textbf{2} & \textbf{Poor} & 
\begin{tabular}[c]{@{}l@{}}
Partially incorrect or confusing.\\ 
Misses critical key steps or includes actions that are irrelevant to the current state,\\ 
making it difficult for the agent to follow.
\end{tabular} \\ 
\midrule
\textbf{1} & \textbf{Critical Failure} & 
\begin{tabular}[c]{@{}l@{}}
Completely hallucinates UI elements that do not exist.\\ 
Logical order is reversed or nonsensical.\\ 
Directly misguides the agent (e.g., "Click Delete" when the goal is "Save").
\end{tabular} \\ 
\bottomrule
\end{tabular}
}
\caption{Human evaluation criteria for generated milestones. The scoring scale ranges from 1 (Critical Failure) to 5 (Perfect), assessing accuracy, feasibility, and granularity.}
\label{tab:rubric_criteria}
\end{table*}
Efficiency is a critical bottleneck in Online Reinforcement Learning, where the agent must interact with the environment and update policies in real-time. To evaluate the computational overhead of our method, we recorded the wall-clock time per epoch over 150 training epochs, comparing ADMIRE against Outcome Reward and Process Reward. The trends are visualized in Figure~\ref{fig:time consumption}.

The results demonstrate that ADMIRE achieves a superior balance between supervision density and computational efficiency. Compared to the lightweight Outcome Reward baseline (Average: 166.83 s/epoch), ADMIRE incurs only a marginal time increase of +12.7\% (Average: 187.99 s/epoch). As shown in the figure, the time consumption curve of ADMIRE closely follows that of the Outcome Reward, indicating that our embedding-based milestone matching mechanism adds negligible latency to the training loop. 

In sharp contrast, the standard Process Reward method is significantly more computationally expensive, likely due to the latency of querying large models for step-by-step verification. It exhibits an average epoch time of 388.44s, representing a substantial +132.8\% increase over the Outcome Reward baseline.

\subsubsection{Milestones Quality}

To ensure that the milestones generated by our dynamic mechanism are reliable and not prone to hallucination, we conducted a human evaluation campaign. We selected the milestones saved after the completion of training and employed human experts to assess the quality of the generated milestones. 

The evaluation follows a 5-point Likert scale, ranging from 1 (Critical Failure) to 5 (Perfect), as defined in Table~\ref{tab:rubric_criteria}. The criteria strictly examine three dimensions: Factuality (Do the described elements exist?), Logical Coherence (Is the order executable?), and Granularity (Is the abstraction level appropriate?).

The results indicate a high level of reliability for ADMIRE. The generated milestones achieved an average score of 4.42, with 87.7\% of the samples rated as 4 (Good) or 5 (Perfect). Cases of hallucination (Score 1) were extremely rare ($<5\%$), primarily occurring in highly ambiguous instruction scenarios. This human verification confirms that the automated milestones generation module produces high-quality milestones that closely align with policy strategy, validating its effectiveness as a supervision signal.

Moreover, this addresses the potential concern regarding objectivity. While milestone generation relies on LLMs, ADMIRE effectively 'anchors' these subjective interpretations into fixed, verifiable text. Unlike the transient and opaque scoring often found in dynamic LLM-as-a-Judge frameworks, our approach solidifies the evaluation criteria into a fixed rubric before reward assignment. Moreover, extracting high-level sub-goals is a comparatively straightforward summarization task for Large Language Models. As evidenced by the evaluation above, LLMs excel at this specific abstraction, ensuring that the resulting reward is not only high-quality but also operationally objective.

\subsection{Milestone Coverage Analysis}
\label{app:milestone_coverage}

\begin{figure}[t]
    \centering
    \includegraphics[width=\linewidth]{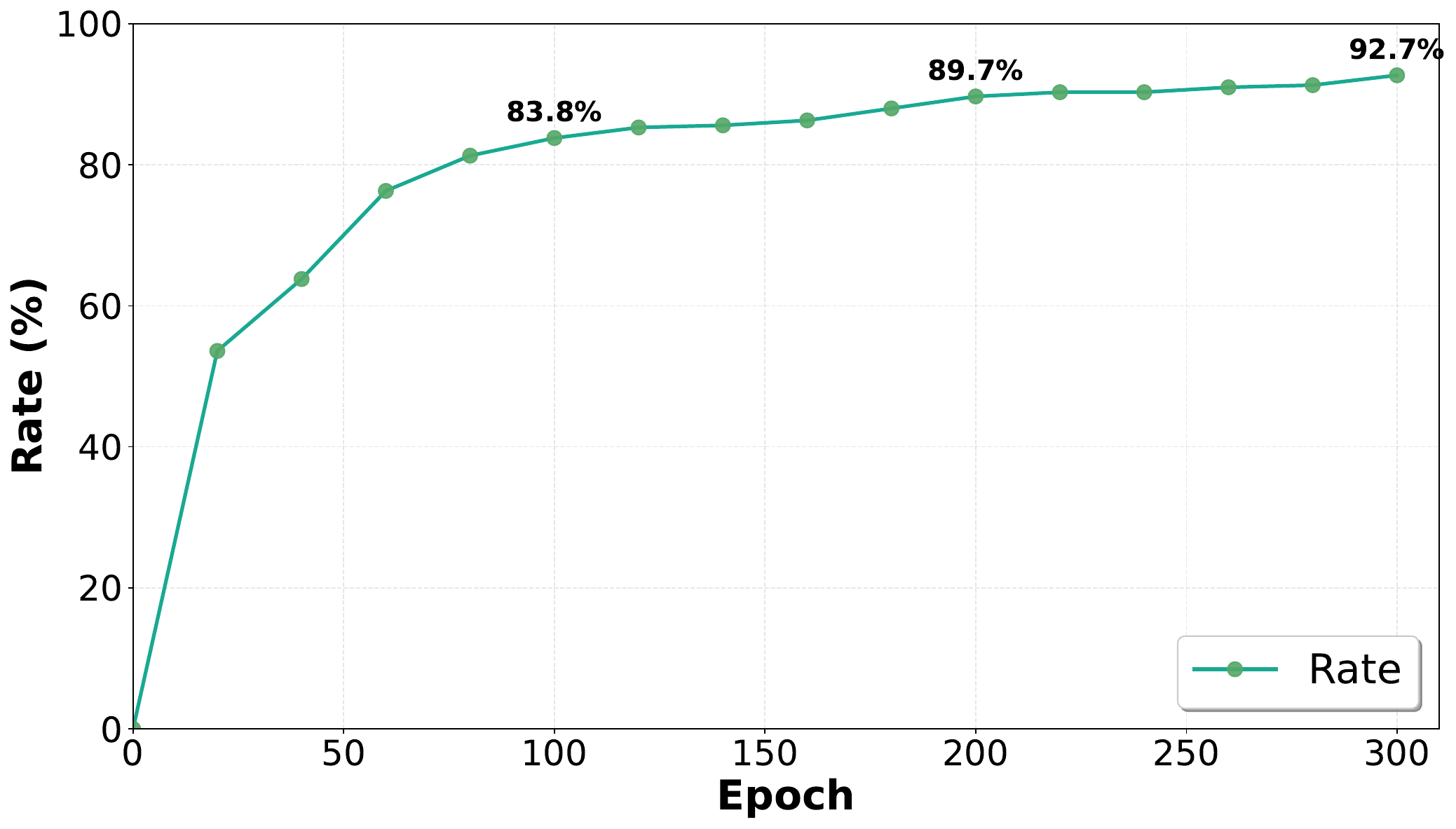}
    \caption{Milestone Initialization Rate over Training Epochs. The coverage of initialized milestones increases rapidly in the early stages and continues to grow steadily, demonstrating that ADMIRE progressively covers 92.7\% of tasks as the agent's capability evolves.}
    \label{fig:milestone_init_rate}
\end{figure}

To understand how the ADMIRE framework adapts throughout the online training process, we analyze the Milestone Initialization Rate, defined as the percentage of tasks for which at least one successful trajectory has been discovered to initialize the milestone memory. As illustrated in Figure~\ref{fig:milestone_init_rate}, the coverage exhibits a rapid and sustained growth pattern. In the initial phase (Epoch 0-100), the rate surges to 83.8\%, indicating that the agent can quickly establish supervision signals for the majority of tasks through early exploration. Subsequently, as the policy optimization proceeds, the agent's capability to handle complex, long-horizon scenarios improves, leading to the unlocking of harder tasks. This results in a steady increase in coverage, which reaches 89.7\% at Epoch 200 and peaks at 92.7\% by the end of training. This trajectory demonstrates that the dynamic generation mechanism effectively co-evolves with the policy, ensuring that dense rewards become available for nearly the entire task spectrum as training deepens.

\subsection{Case Study}
\label{app:case}
\begin{figure*}[ht]
    \centering
    \includegraphics[width=\textwidth]{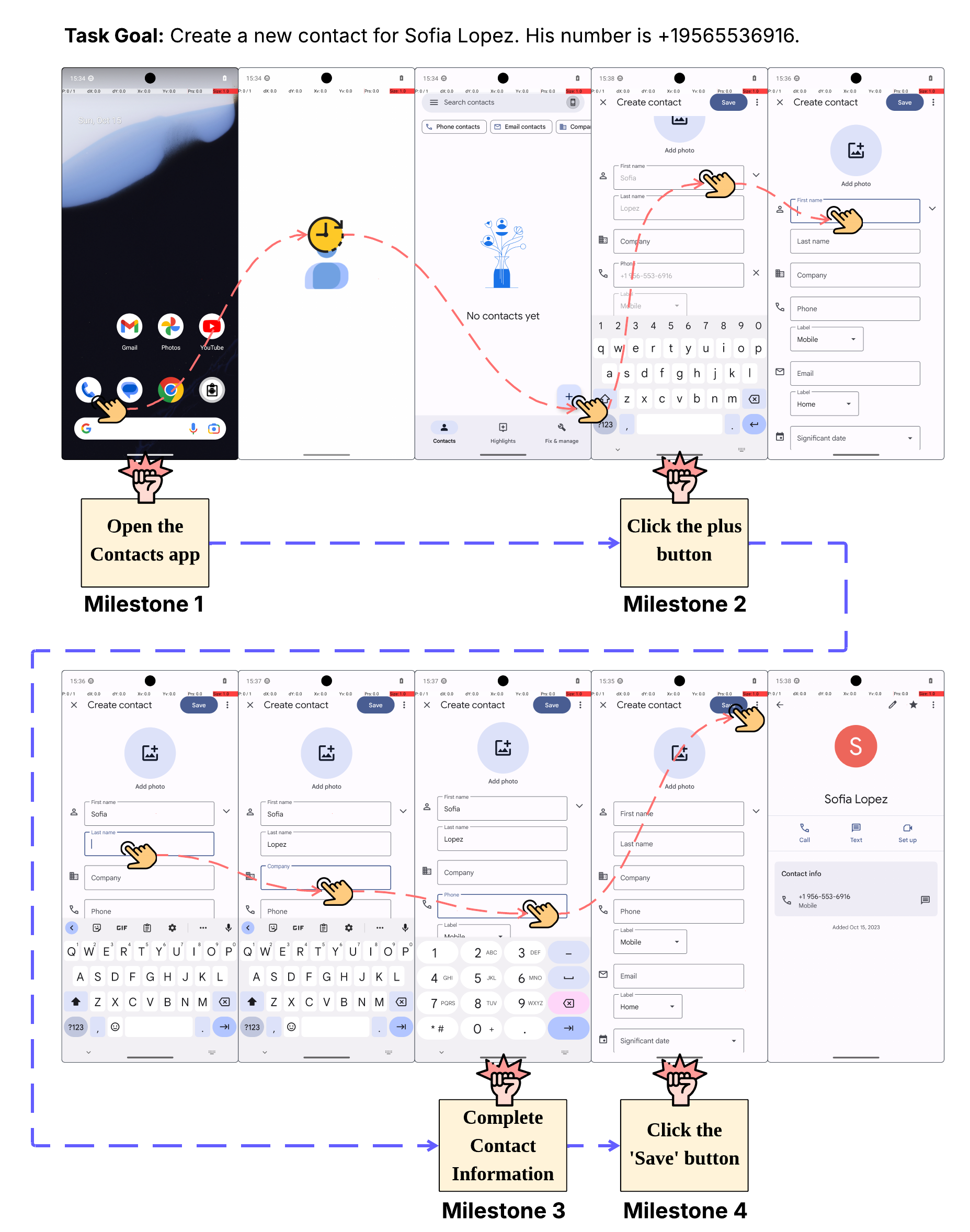}
    \caption{A Case Study on Milestone Alignment for Mobile GUI Agent Task Trajectories}
    \label{fig:case success}
\end{figure*}
Figure~\ref{fig:case success} presents a case study showing how milestones match the trajectories generated by the Mobile GUI agent. The figure shows that milestones are matched based on the agent's progress, offering more comprehensive guidance to the agent in the online RL process.

\section{Large Language Models (LLMs) Usage}
In this work, large language models (LLMs) were used solely for text polishing and language refinement. They were not involved in the design of the methodology, implementation, analysis, or generation of experimental results. All technical contributions and research findings are entirely the work of the authors.

\begin{figure*}[t]
\centering
\begin{tcolorbox}[colback=yellow!10!white, colframe=gray!80!black, title=Prompt for Milestone–Action Description Matching]
You are an expert milestone judge. 

Sentence A is a milestone — it describes a required step for completing a task. 
Sentence B is the current action description executed by an agent.

Your job is to determine whether Sentence B successfully fulfills (matches) the milestone described in Sentence A.

Follow these guidelines strictly:

* Respond "yes" **only** if Sentence B clearly accomplishes the same intent and target as the milestone. 

* Answer "no" when the action targets something else, represents a different gesture, only partially covers the milestone, or leaves uncertainty about completion.

* Differences in wording, phrasing, or word order **do not matter** — focus solely on whether the **core meaning** of the two sentences aligns.

* If the main goal ultimately achieved is the same, then even with minor differences, the two sentences should be considered a match and Answer "yes".

* If Sentence B encompasses Sentence A, or if the action in Sentence B is a stronger version of that in Sentence A, they should also be considered a match and Answer "yes".

* If Sentence B is a more specific version of Sentence A, they should also be considered a match and Answer "yes".

* <...> is a placeholder for any replaceable component in the milestone, such as <App>, <Action>, <Target>, <Direction>, <Value>, etc. Any component that can be replaced by <...> and matches the milestone should be considered as a hit and Answer "yes".    

Respond strictly with lowercase "yes" or "no" without any explanation.

\noindent \\
Milestone (Sentence A):\{\textit{Milestone}\}

Current action description (Sentence B):\{\textit{Action\_Description}\}

Question: Does the current action hit the milestone? **Answer 'yes' or 'no' only.**
\end{tcolorbox}
\caption{Prompt used to determine whether a milestone matches an action description.}
\label{fig:Prompt match}
\end{figure*}

\begin{figure*}[t]
\centering
\begin{tcolorbox}[colback=yellow!10!white, colframe=gray!80!black, title=Prompt for Initializing Task Milestones]
You are an expert in mobile GUI operations. You will be given a successfully completed task trajectory, including:

* the task name and overall goal,

* a sequence of action descriptions representing user interactions, and

* corresponding reference screenshots.

Your task is to identify only the **essential steps** required to accomplish the goal. Focus exclusively on actions that are absolutely necessary, and **ignore redundant or non-essential steps**.

Your objective is to distill these essential actions into a **concise list of milestones items** representing the key sub-tasks needed to complete the goal. These milestones:

* should represent **logically distinct stages** of progress toward the goal,

* must be **clear**,

* should be **short, non-overlapping, and non-redundant**, and

* must include **only necessary steps**.

* Keep important action-related keywords unchanged.

When task-specific details appear (such as names, dates, events, entries, files, or paths), replace them with generic placeholders like `<Name>`, `<Date>`, `<Event>`, `<Entry>`, `<File>`, or `<Path>`.

Your final output must be a **strict JSON array** containing only these milestone strings.

Do **not** include any numbering, explanations, commentary, or other non-JSON content.

Focus on summarizing the **core actions** into a clean, abstract milestone — generalizable and free of specific details.

For example: "Locate the <Entry> on the page", "Input <Name> as the note name", "Navigate to the <Path> on the page".

\noindent \\
Task: \{\textit{task}\}

User Instruction: \{\textit{Instruction}\}

Example action descriptions from new trajectory (ordered; may be redundant):\{\textit{Action Description List}\}

Output the essential milestone as a strict JSON array.

\end{tcolorbox}
\caption{Prompt used to initialize the task's milestones based on the correct trajectory.}
\label{fig:Prompt initialize}
\end{figure*}

\begin{figure*}[t]
\centering
\begin{tcolorbox}[colback=yellow!10!white, colframe=gray!80!black, title=Prompt for Refining Task Milestones]
You are an expert in refining mobile GUI task milestones. You will be provided with the following information:

* the task name and overall goal,

* an existing milestone,

* a set of action descriptions from a newly observed successful trajectory, and

* corresponding reference screenshots.

Your role is to analyze the new trajectory, filtering out any redundant or meaningless interactions and focusing only on actions that are **essential** for achieving the task's objective.

Your primary responsibility is to determine whether the **existing milestone** can be improved based on the new information.

By default, assume the current milestones are already sufficient. They should be revised **only** if the new trajectory clearly reveals **more minimal, precise, or effective** actions that better represent the essential steps.

If the existing milestone already provides a clear, general, and minimal set of essential steps, return it **unchanged**.

If an update is warranted, output the improved milestones as a **strict JSON array**.
Each milestone must be:

* a short, **executable step** starting with an action verb,

* **non-redundant** and **clearly phrased**, and

* generalized by replacing task-specific details (e.g., names, dates, events, entries, paths) with placeholders such as `<Name>`, `<Date>`, `<Event>`, `<Entry>`, or `<Path>`.

Do **not** include any numbering, commentary, or text outside the JSON structure.

Your output must contain **only** the finalized milestones in valid JSON format.

\noindent \\
Task: \{\textit{task}\}

User Instruction: \{\textit{Instruction}\}

Existing milestone (JSON):\{\textit{Milestone}\}

Example action descriptions from new trajectory (ordered; may be redundant):\{\textit{Action Description List}\}

Decide whether to keep the existing milestone or refine it. Output ONLY the final milestone as a strict JSON array.
\end{tcolorbox}
\caption{Prompt used to refining the task's milestones based on the correct trajectory.}
\label{fig:Prompt Refine}
\end{figure*}

\begin{figure*}[t]
\centering
\begin{tcolorbox}[colback=yellow!10!white, colframe=gray!80!black, title=Prompt for LLM-as-Judge]
You are a step helpfulness judge with precise evaluative capabilities: you can accurately determine whether a task has been completed, assess the relevance of actions to goals, and distinguish between meaningful progress and irrelevant actions. Your task is to evaluate a single-step trajectory of a GUI agent executing a task and determine whether this specific step is helpful for achieving the goal. 

* What yor are provided
You are given a mobile GUI task goal, the current step's action text (including the model's thought process and output action), and the corresponding screenshots before and after the action. Red dots on the screenshots indicate click positions, while red arrows represent scroll operations and their directions. You are required to determine whether this step (action) helps progress toward completing the goal.

* If the task is completed at this step, its status is 'success', or the action is 'terminate', return 'yes' directly.

* If a discrepancy is found between the model's thought process and its final action, the action shall serve as the primary basis for judgment.

* Helpful steps include, but are not limited to: entering relevant information, advancing to the next stage, selecting target options, or saving progress.

* Unhelpful steps include: irrelevant taps, dismissing relevant dialogs, opening unrelated or wrong apps, idle, incomplete or wrong operations and so on.

* If there is uncertainty about whether it is helpful or the situation is ambiguous, Answer "no".

* If the action description or action text is empty or None, you should give 'no' directly.

The following are some examples for your reference:

Example for **helpful steps**:

Task goal: Fill out an expense report.
Action text: "Tap on the 'Add' button."
Action description: "Click the 'Add' button at the top right corner of the page."
Screenshot: A screenshot showing a button labeled 'Add' at the top right corner of the page  while a 'delete' button is on the left side of the page. The red dot is on the 'Add' button.
Answer: "yes"

Example for **unhelpful steps**:

Task goal: Fill out an expense report.
Action text: "Tap on the 'Add' button."
Action description: "Click the 'Add' button at the top right corner of the page."
Screenshot: A screenshot showing a button labeled 'Add' at the top right corner of the page while a 'delete' button is on the left side of the page. But the red dot is on the 'delete' button.
Answer: "no"

The helpfulness of a step is not only determined by the action text and description, but also the screenshot. You should only give 'yes' for those steps that demonstrate clear and meaningful progress toward the goal. Be very strict and conservative - only give 'yes' when you are absolutely certain the step directly contributes to completing the task. If there is any doubt or uncertainty about whether a step is truly helpful, you must give 'no'.

Your should **Answer strictly with 'yes' or 'no'** (lowercase) directly, without any explanation.

\noindent \\
Goal: \{\textit{goal}\}

Step Action Text:\{\textit{action}\}

Step Action Description: \{\textit{action description}\}

Question: Does this step help progress toward the goal? Answer 'yes' or 'no' directly.
\end{tcolorbox}
\caption{Prompt Used for LLM-Based Process Reward Judging.}
\label{fig:Prompt PRM}
\end{figure*}

\begin{figure*}[t]
\centering
\begin{tcolorbox}[
  enhanced,
  colback=yellow!10!white,
  colframe=gray!80!black,
  title={Prompt for GUI Agent (Part I)}
]
You are an agent who can operate an Android phone on behalf of a user. Based on user's goal/request, you may

- Answer back if the request/goal is a question (or a chat message), like user asks ``What is my schedule for today?''.

- Complete some tasks described in the requests/goals by performing actions (step by step) on the phone.

When given a user request, you will try to complete it step by step. At each step, you will be given the current screenshot (the original screenshot) and a history of what you have done (in text). Based on these pieces of information and the goal, you must choose to perform one of the action in the following list (action description followed by the JSON format) by outputing the action in the correct JSON format.

- If you think the task has been completed, finish the task by using the status action with complete as goal status:  
  `{{"action type": "status", "goal status": "complete"}}`

- If you think the task is not feasible (including cases like you don't have enough information or can not perform some necessary actions), finish by using the `status` action with infeasible as goal status:  
  `{{"action type": "status", "goal status": "infeasible"}}`

- Answer user's question:  
  `{{"action type": "answer", "text": "<answer text>"}}`

- Click/tap on an element on the screen:  
  `{{"action type": "click", "x": <target x>, "y": <target y>}}`.

- Long press on an element on the screen:  
  `{{"action type": "long press", "x": <target x>, "y": <target y>}}`.

- Type text into a text field:  
  `{{"action type": "input text", "text": <text input>, "x": <target x>, "y": <target y>}}`

- Press the Enter key:  
  `{{"action type": "keyboard enter"}}`

- Navigate to the home screen:  
  `{{"action type": "navigate home"}}`

- Navigate back:  
  `{{"action type": "navigate back"}}`

- Scroll the screen or a scrollable UI element:  
  `{{"action type": "scroll", "direction": <up, down, left, right>, "x": <optional target x>, "y": <optional target y>}}`

- Open an app:  
  `{{"action type": "open app", "app name": <name>}}`

- Wait for the screen to update:  
  `{{"action type": "wait"}}`

The current user goal/request is: \{goal\}

Here is a history of what you have done so far:  
\{history\}

The current screenshot is also given to you.
\end{tcolorbox}

\caption{Prompt for GUI agent during training (Part I: task definition, action space, and execution context).}
\label{fig:prompt_gui_part1}
\end{figure*}

\begin{figure*}[t]
\centering
\begin{tcolorbox}[
  enhanced,
  colback=yellow!10!white,
  colframe=gray!80!black,
  title={Prompt for GUI Agent (Part II)}
]
Here are some useful guidelines you need to follow:

\textbf{General:}

- Usually there will be multiple ways to complete a task, pick the easiest one. Also when something does not work as expected (due to various reasons), sometimes a simple retry can solve the problem, but if it doesn't (you can see that from the history), SWITCH to other solutions.

- Sometimes you may need to navigate the phone to gather information needed to complete the task, for example if user asks ``what is my schedule tomorrow'', then you may want to open the calendar app (using the `open app` action), look up information there, answer user's question (using the `answer` action) and finish (using the `status` action with complete as goal status).

- For requests that are questions (or chat messages), remember to use the `answer` action to reply to user explicitly before finish! Merely displaying the answer on the screen is NOT sufficient (unless the goal is something like ``show me ...'').

- If the desired state is already achieved (e.g., enabling Wi-Fi when it's already on), you can just complete the task.

\medskip
\textbf{Action Related:}

- Use the `open app` action whenever you want to open an app (nothing will happen if the app is not installed), do not use the app drawer to open an app unless all other ways have failed.

- Use the `input text` action whenever you want to type something (including password) instead of clicking characters on the keyboard one by one. Sometimes there is some default text in the text field you want to type in, remember to delete them before typing.

- For `click`, `long press` and `input text`, the index parameter you pick must be VISIBLE in the screenshot and also in the UI element list given to you (some elements in the list may NOT be visible on the screen so you can not interact with them).

- Consider exploring the screen by using the `scroll` action with different directions to reveal additional content.

- The direction parameter for the `scroll` action can be confusing sometimes as it's opposite to swipe, for example, to view content at the bottom, the `scroll` direction should be set to ``down''. It has been observed that you have difficulties in choosing the correct direction, so if one does not work, try the opposite as well.
\end{tcolorbox}

\caption{Prompt for GUI agent during training (Part II: general principles and action-level guidelines).}
\label{fig:prompt_gui_part2}
\end{figure*}

\begin{figure*}[t]
\centering
\begin{tcolorbox}[
  enhanced,
  colback=yellow!10!white,
  colframe=gray!80!black,
  title={Prompt for GUI Agent (Part III)}
]
\textbf{Text Related Operations:}

- Normally to select certain text on the screen:  
  \textit{(i)} Enter text selection mode by long pressing the area where the text is, then some of the words near the long press point will be selected (highlighted with two pointers indicating the range) and usually a text selection bar will also appear with options like `copy`, `paste`, `select all`, etc.  
  \textit{(ii)} Select the exact text you need. Usually the text selected from the previous step is NOT the one you want, you need to adjust the range by dragging the two pointers. If you want to select all text in the text field, simply click the `select all` button in the bar.

- At this point, you don't have the ability to drag something around the screen, so in general you can not select arbitrary text.

- To delete some text: the most traditional way is to place the cursor at the right place and use the backspace button in the keyboard to delete the characters one by one (can long press the backspace to accelerate if there are many to delete). Another approach is to first select the text you want to delete, then click the backspace button in the keyboard.

- To copy some text: first select the exact text you want to copy, which usually also brings up the text selection bar, then click the `copy` button in bar.

- To paste text into a text box, first long press the text box, then usually the text selection bar will appear with a `paste` button in it.

- When typing into a text field, sometimes an auto-complete dropdown list will appear. This usually indicating this is a enum field and you should try to select the best match by clicking the corresponding one in the list.

\medskip
Now output an action from the above list in the correct JSON format, following the reason why you do that. Your answer should look like:

\medskip
Reason: ...

Action: \{\{"action type":...\}\}

\medskip
Your Answer:
\end{tcolorbox}

\caption{Prompt for GUI agent during training (Part III: text manipulation rules and output format).}
\label{fig:prompt_gui_part3}
\end{figure*}